\documentclass[11pt,a4paper]{article}
\usepackage[hyperref]{acl2020}
\usepackage{times}
\usepackage{latexsym}
\usepackage{footmisc}

\usepackage{microtype}

\usepackage{times}
\usepackage{latexsym}
\usepackage{url}
\usepackage{bbm}
\usepackage[T1]{fontenc}
\usepackage{graphicx}
\usepackage{amsmath}

\usepackage{url}
\usepackage{comment}

\aclfinalcopy 

\newcommand\csr{\textsc{CsrQA}}
\newcommand\rsr{\textsc{RsrQA}}
\newcommand\cs{\textsc{CsQA}}
\newcommand\crq{\textsc{CrQA}}
\title{Large Scale Question Answering using Tourism Data}
\author{Danish Contractor${^1}{^,}$${^2}$\thanks{This work was carried out as part of PhD research at IIT Delhi.The author is also a regular employee at IBM Research.} \quad Krunal Shah$^3$\thanks{Work carried out when the author was a student at IIT Delhi.} \quad Aditi Partap${^4}{^\dagger}$ \quad Mausam$^2$ \quad Parag Singla$^2$\\
$^1$IBM Research AI, New Delhi\qquad $^2$Indian Institute of Technology, New Delhi \\$^3$University of Pennsylvania, Philadelphia
\qquad 
$^4$University of Illinois, Urbana Champaign\\ \\
\texttt{dcontrac@in.ibm.com, shahkr@seas.upenn.edu , aaditi2@illinois.edu}, \\  \{\texttt{mausam,parags\}@cse.iitd.ac.in}}
 
\begin{document}

\maketitle

\begin{abstract}

We introduce the novel task of answering entity-seeking recommendation questions using a collection of reviews that describe candidate answer entities. We harvest a QA dataset that contains 47,124 paragraph-sized real user questions  from  travelers  seeking  recommendations for hotels, attractions and restaurants. Each question can have thousands of candidate answers to choose from and each candidate is associated with a collection  of unstructured  reviews. This dataset is especially challenging because commonly used neural architectures for reasoning and QA are prohibitively expensive for a task of this scale. As a solution, we design a scalable cluster-select-rerank approach. It first clusters text for each entity to identify exemplar sentences describing an entity. It then uses a scalable neural information retrieval (IR) module to select a set of potential entities from the large candidate set. A reranker uses a deeper attention-based architecture to pick the best answers from the selected entities. This strategy performs better than a pure IR or a pure attention-based reasoning approach yielding nearly 25\% relative improvement in Accuracy@3 over both approaches.

\end{abstract}

\section{Introduction} \label{sec:introduction}
\begin{figure*}[h]
 {\footnotesize
 \center
   \includegraphics[scale=0.41]{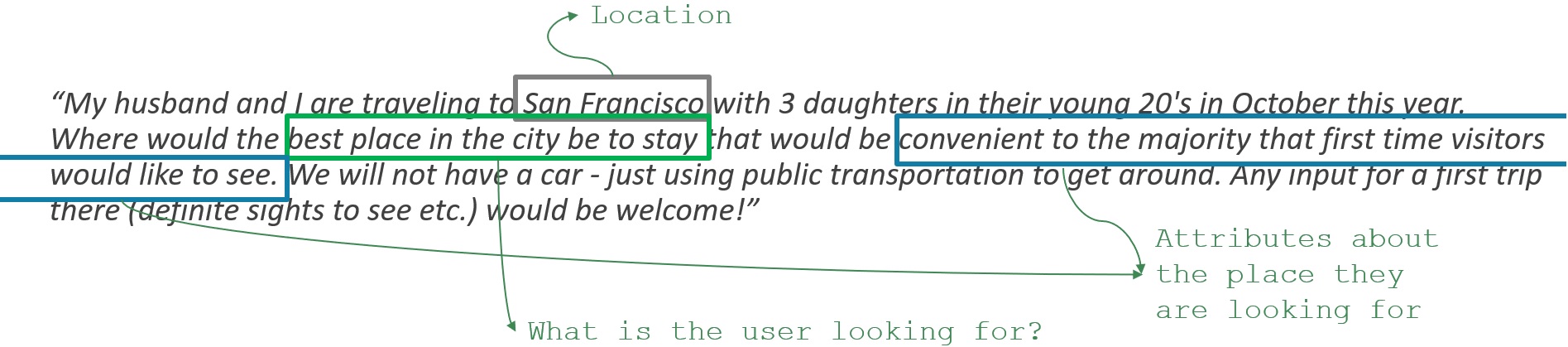}
   \caption{\small Question from an online travel forum. Figure adapted from \cite{MSEQ2}}
   \label{fig:queryExample}
 
 }
\vspace*{-3ex}
 \end{figure*}
Real-world questions, such as those seen on online forums are often verbose, requiring us to first determine what is crucial in the question for answering. For example, consider the question in Figure \ref{fig:queryExample}. Here the user describes {\em who they are}, {\em what they are looking for}, as well as 
{\em their preferences} for the expected answer.
They also mention that they are {\em looking forward to the trip} and are going to be {\em first time visitors} to the city.  
Answering such questions requires understanding the relevant parts of the question, 
reading information about each candidate answer entity in travel articles, blogs or reviews (\emph{entity documents}), matching relevant question parts with entity documents, and ranking each candidate answer based on the degree of match. 

In this paper we introduce the
novel task of answering such entity-seeking recommendation questions using a collection of reviews describing entities. Our task reflects real-word challenges of reasoning and scale. 

\noindent{\bf Reasoning:}
Entity reviews written by users can be informal and noisy, and may contain subjective contradictory opinions. Further, reviews may also discuss other entities (e.g., for comparison), making reasoning even harder. Finally, not all aspects of the question are relevant for answering which makes identifying the informational need challenging. It is worth noting that the question in Figure \ref{fig:queryExample} is almost as large as the reading comprehension paragraphs used in tasks such as SQuAD.\footnote{Average size of paragraphs in SQuAD is 140 tokens}

\noindent {\bf Scalability}: Typical QA algorithms apply cross-attention between question and candidate answer texts, which do not scale in our task where entities may have long review documents (see Table \ref{tab:related} for comparison of document sizes across different QA datasets).  Moreover, the candidate answer spaces for this problem are very high (e.g., New York has tens of thousands of restaurants to choose from), affecting scalability further.

\subsection{Contributions}\label{sec:contrib}
 We introduce the novel task of answering entity-seeking recommendation questions using a collection of reviews describing entities. Formally, given an entity seeking recommendation question ($q$), its target class ($t \in $ \{hotel, attraction, restaurant\}), the city $c$, a candidate space of entities $E_c^t$ for each corresponding city and entity class, a collection of their reviews $R_c^t$; the goal of this task is to find the correct (gold) answer entity $a \in E_c^t$ using the documents $R_c^t$ describing the candidate answer entities. It is inspired by the recent work on parsing  multi-sentence  questions \cite{MSEQ2}. Our work differs from theirs, because they do not attempt to solve this QA task end-to-end and, instead, rely on a pipeline of semantic parsing and querying into a Web API. In contrast, we harvest a novel dataset \footnote{We release scripts to regenerate the dataset.} of tourism questions consisting of $47,124$ QA pairs extracted from online travel forums. Each QA pair consists of a question and an answer entity ID, which corresponds to one of the over $200,000$ entity review documents collected from the Web. 
The entities in our dataset are
hotels, restaurants and general attractions of interest in $50$ cities across the world. 
Gold-answer entities are extracted from mentions in full text user responses to the actual forum post. To the best of our knowledge, we are the first to propose and attempt such an end-to-end task using a collection of entity reviews.

In addition to a QA task, our task can also be viewed as an instance of  information retrieval (IR), since we wish to return entity documents (equivalent to returning entities), except that the query is a long question. IR models are more scalable, as they often have methods aimed at primarily matching and aggregating information \cite{IRSurvey}. Thus, these models typically do not achieve deep reasoning, which QA models do, but as mentioned previously, deeper reasoning in QA models (eg. using multiple layers of cross attention) makes them less scalable. We therefore, propose a Cluster-Select-Rerank (\csr) architecture for our task, drawing from strengths of both.

The $\csr$ architecture first {\em clusters} text for each entity to identify exemplar sentences describing an entity. It then uses a scalable neural information retrieval (IR) module to {\em select} a set of potential entities from the large candidate set. A {\em reranker} uses a deeper attention-based architecture to pick the best answers from the selected entities. This strategy performs better than a pure IR or a pure attention-based reasoning approach yielding nearly 25\% relative improvement in Accuracy@3 over both approaches.
\section{Related Work} \label{sec:related}
\begin{table*}
\scriptsize
\center
\begin{tabular}{c|c|c|c|c|c}
\textbf{Dataset} & \multicolumn{1}{c}{\textbf{\begin{tabular}[c]{@{}c@{}}Knowledge\\ Source\end{tabular}}} & \multicolumn{1}{c}{\textbf{\begin{tabular}[c]{@{}c@{}}Answer \\ type\end{tabular}}} & \multicolumn{1}{c}{\textbf{\begin{tabular}[c]{@{}c@{}}Avg. tokens\\ in documents\end{tabular}}} & \multicolumn{1}{c}{\textbf{\begin{tabular}[c]{@{}c@{}}Answer \\ document\\ known\end{tabular}}} & \multicolumn{1}{c}{\textbf{\begin{tabular}[c]{@{}c@{}}Multiple docs* \\ required\\ for answering\end{tabular}}}  \\ \hline \hline
\textbf{SQuAD} \cite{SqUAD}        & Wikipedia paragraphs                                                                                         &  Span                                                                                    &  137                                                                                               & Y  & N                                                                                             \\
\textbf{NewsQA}\cite{NewsQA}        & CNN News articles                                                                                        &   Span                                                                                  &  $\approx$ 300                                                                                               & Y  & N                                                                                               \\
\textbf{SearchQA} \cite{SearchQA}       & Web snippets                                                                                        &  Span                                                                                   &  25-46                                                                                               &    N & N                                                                                           \\
\textbf{RACE} \cite{RACE}       & Passages on topics                                                                                        &  Choices                                                                                & 350                                                                                                 & Y               &  N                                                                                \\
\textbf{OpenBookQA} \cite{OpenBookQA2018}       & Fact sentences                                                                                        &  Choices                                                                                &   10                                                                                             & Y               &  Y                                                                                \\
\textbf{MSMARCO} \cite{MSMARCO}       &   Web article snippets                                                                                     &  Free text                                                                                   & 10                                                                                                & N         & Y                                                                                      \\
\textbf{MedQA} \cite{MedQA}        & Medical Articles                                                                                        & Choices                                                                                    &  35                                                                                               & Y              & Y                                                                                 \\
\textbf{WikiReading} \cite{WikiReading}        &  Wikipedia articles                                                                                      & Infobox property                                                                                    &                                                                                         489       &    N                  &    N                                                                       \\
\textbf{TriviaQA} \cite{TriviaQA}        &  Web articles                                                                                       &  Span                                                                                   & 2895$^{\dagger}$                                                                                       & N & Y \\
\textbf{HotPot-QA} \cite{HotpotQA}        & Wikipedia paragraphs                                                                                         & Span                                                                                    &        $\approx$ 800                                                                                         &  N                                                                & Y                               \\
\textbf{ELI5} \cite{LongFormQA}         &  Passages on topics                                                                                       &    Free text                                                                              &                                                                                 858                &     Y                                                                       &    N                \\
\textbf{TechQA} \cite{TechQA}         &  IT support notes                                                                                       &    Free text                                                                              &                                                                                 48                &     Y                                                                       &    Y                \\

\hline
\hline

Dialog QA - \textbf{QuAC} \cite{QuAC}         &  Wikipedia passages                                                                                       &    Span                                                                              &                                                                                               401  &     Y                                                                        &    N                \\ 

Dialog QA - \textbf{CoQA} \cite{CoQA}         &  Passages on topics                                                                                       &    Free text                                                                              &                                                                                 271                &     Y                                                                       &    N                \\

\textbf{Our Dataset}       &  Reviews                                                                                       & Entity (doc.)                                                                                    &                                                                                  3266 & N & Y 
\end{tabular}
\vspace*{-1ex}
\caption{\small Related datasets on Machine reading/QA and their characteristics. For reading comprehension tasks, the document containing the actual answer may not always be known. *"docs" refers to what the task would consider as its document (e.g., fact sentences for OpenBookQA). $^{\dagger}$Most questions in TriviaQA are answerable using only the first few hundred tokens in the document.} 
\label{tab:related}
\vspace*{-4ex}
\end{table*}

\noindent{\bf QA Tasks: }Recent question answering tasks such as those based on reading comprehension require answers to be generated either based on a single passage, or after reasoning over multiple passages (or small-sized documents) (e.g. {\em SQuAD} \cite{SqUAD}, {\em HotpotQA} \cite{HotpotQA}, {\em NewsQA} \cite{NewsQA}). Answers to questions are assumed to be stated explicitly in the documents \cite{SqUAD} and can be derived with single or multi-hop reasoning over sentences mentioning facts \cite{HotpotQA}. Other variants of these tasks add an additional layer of complexity where the document containing the answer may not be known and needs to be retrieved from a large corpus before answers can be extracted/generated (e.g. {\em SearchQA} \cite{SearchQA}, {\em MS MARCO} \cite{MSMARCO}, {\em TriviaQA} \cite{TriviaQA}). Models for these tasks typically use variants of TF-IDF like BM25 ranking \cite{BM25} to retrieve and sub-select candidate documents \cite{drqa}; reasoning is then performed over this reduced space to return answers.
However, we find that in our task retrieval strategies such as BM25 perform poorly\footnote{Accuracy@3 of 7\%} and are thus not effective in reducing the candidate space (see Section \ref{sec:experiments}). As a result, our task requires processing $500$ times more documents per question and also requires reasoning over large entity review-documents (Table \ref{tab:related}) that consist of noisy, subjective opinions.  Further, traditional QA models such as BiDAF \cite{BIDAF} or those based on BERT \cite{BERT} are infeasible\footnote{BiDAF requires 43 hours for 1 epoch (4 K-80 GPUs)} to train for our task. Thus, while existing tasks and datasets have been useful in furthering research in comprehension, inference and reasoning, we find that they do not always reflect 
all the complexities of real-world question answering motivated in our task.



\noindent{\bf IR Tasks: }Our QA task is one that also shares characteristics of information retrieval (IR), because, similar to document retrieval, answers in our task are associated with long entity documents, though they are without any additional structure. The goal of IR, specifically document retrieval tasks, is to retrieve documents for a given query. 
Typical queries in these tasks are short, though some IR works have also studied long queries \cite{LiveQA}. Documents in such collections tend to be larger than passages and often retain structure -- titles, headings, etc. Neural models for IR focus on identifying good representations for queries and documents to maximize mutual relevance in latent space \cite{IRSurvey}. To improve dealing with rare words recent neural models also incorporate lexical matching along with semantic matching \cite{DUET}. However, unlike typical retrieval tasks, the challenge for answering in our task is not merely that of semantic gap -- subjective opinions need to be {\em reasoned} over and aggregated in order to assess relevance of the entity document. This is similar to other reading comprehension style QA tasks that require deeper reasoning over text.

We believe that this setting brings together an interesting mix: (i) a large search space with large documents (like in IR), and that (ii) answering cannot rely only on methods that are purely based on semantic and lexical overlap (it requires reasoning). 
Thus, in this paper we present a coarse-to-fine algorithm that sub-selects documents using IR and trains a deep reasoner over the selected subset (Section \ref{sec:model}).

\section{Data Collection} \label{sec:data-collection}
Most recent QA datasets have been constructed using crowdsourced workers who either create QA pairs given documents~\cite{SqUAD,CoQA} or identify answers for real world questions~\cite{MSMARCO,NaturalQuestions}. Creating QA datasets manually using the crowd can be very expensive and we therefore choose to automatically harvest a dataset using forums and a collection of reviews. We first crawled forum posts along with their corresponding conversation thread as well as meta-data including date and time of posting. We then also crawled reviews for restaurants and attractions for each city from a popular travel forum. Hotel reviews were scraped from a popular hotel booking website. 
Entity meta-data such as the address, ratings, amenities, etc was also collected where available. 

We observed that apart from questions, forum users 
%
also 
post summaries of trips, feedback about services taken during a vacation, open-ended non entity-seeking questions such as queries about the weather and economic climate of a location, etc. Such questions were removed by precision oriented rules which discarded questions that did not contain any one of the phrases in the set [``{\it recommend}'', ``{\it suggest}'', ``{\it where}'', ``{\it place to}'' ``{\it best''} and ``{\it option}'']. While the use of such rules may introduce a bias towards a certain class of questions, as Table \ref{tab:questions} suggests,  they continue to retain a lot of variability in language of expression that still makes the task challenging. We further removed posts explicitly identified as ``Trip Reports'' or ``Inappropriate" by the forum. Excessively long questions ($\geq$ 1.7X more than average) were also removed. 

\subsection{Answer Extraction}
We create a list of entity names crawled for each city and use it to find entity mentions in user responses to forum posts. A high level entity class (hotel, restaurant, attraction) for each entity is also tagged based on the source of the crawl. Each user response to a question is tagged for part-of-speech, and the nouns identified are fuzzily searched\footnote{Levenstein distance<0.05} in the entity list (to accommodate for typographical errors). This gives us a noisy set of ``{\em silver}" answer entities extracted from free text user responses for each question. We now describe a series of steps aimed at improving the precision of extracted silver answers, resulting in our gold QA pairs.

\subsection{Filtering of Silver Answer Entities} \label{sec:filtering}

{\bf Question Parsing:} As a first step, we use the multi-sentence question understanding component developed by \citeauthor{MSEQ2} (\citeyear{MSEQ2}) to identify phrases in the question that could indicate a target entity's ``{\em type}'' and ``{\em attribute}''. For instance, in the example in Figure \ref{fig:queryExample} tokens ``{\em place to stay}'' will be identified as an $entity.type$ while ``{\em convenient to the majority of first time visitors}'' will be identified as $entity.attribute$. 

\noindent {\bf Type-based filtering:} All entities collected from the online forums come with labels (from a set of nearly 210 unique labels) indicating the nature of the entity. For instance, restaurants have cuisine types mentioned, attractions are tagged as museums, parks etc. Hotels from the hotel booking website are simply identified as ``hotels''. 
We manually cluster the set of unique labels into 11 clusters. 
For a given question we use the phrase tagged with the $entity.type$ label from the question parse, and determine its closest matching cluster using embedding representations. Similarly, for each silver answer entity extracted we identify the most likely cluster given its tagged attribute list; if the two clusters do not match, we drop the QA pair. 


\noindent {\bf Peer-based filtering:} As mentioned previously, all entities and their reviews contain labels in their meta-data indicating the nature of the entity. Using all {\em silver} (entity) answers for a question, we determine the frequency counts of each label encountered (an entity can be labeled with more than one label by the online forum). We then compare the label of each {\em silver} answer with the most frequent label and remove any silver (entity) answer that does not belong to the majority label.

\noindent{\bf Filtering entities with generic names:} Some entities are often named after cities, or generic place types  -- for example ``The Cafe'' or ``The Spa'' which can result in spurious matches during answer extraction. We collect a list of entity types\footnote{Examples of types include ``cafe'', ``hospital'', ``bar'' etc.} from Google Places\footnote{\emph{https://developers.google.com/places/web-service/supported\_types}} and remove any answer entity whose name matches any entry in this list. 

\noindent{\bf Removing entities that are chains and franchises:} Answers to questions can also be names of restaurant or hotel chains without adequate information to identify the actual franchisee referred. In such cases, our answer extraction returns all entities in the city with that name. We thus, discard all such QA pairs. 

\noindent{\bf Removing spurious candidates:} User answers in forum posts often have multiple entities mentioned not necessarily in the context of an answer but for locative references (e.g. ``opposite Starbucks'', or ``near Wendys'') or for expressing opinions on entities that are not the answer. We write simple rules to remove candidates extracted in such conditions (e.g.: if more than one entity is extracted from a sentence, we drop them all or if entity mentions are in close proximity to phrases such as ``next to'', ``opposite'' etc. they are dropped). 

Additionally, we review the set of entities extracted and remove QA pairs with entity names that were common English words or phrases (eg: ``August'', ``Upstairs'', ``Neighborhood'' were all names of restaurants that could lead to spurious matches). We remove $322$ unique entity names as a result of this exercise. Note that it is the only step that involved human annotation in the data collection pipeline thus far.

 \subsection{Crowd-sourced Data Cleaning}
We expect that our automated QA pair extraction methods are likely to have some degree of noise (See Section \ref{sec:qual}). In order to facilitate accurate bench-marking, we crowd-source and clean our validation and test sets. 
We use the Amazon Mechanical Turk(AMT)\footnote{http://requester.mturk.com} for crowd-sourcing\footnote{See Supplementary Notes}. Workers are presented with a QA-pair, which includes the original question, an answer-entity extracted by our rules and the original forum-post response thread where the answer entity was mentioned. Workers are then asked to check if the extracted answer entity was mentioned in the forum responses as an answer to the user question. We spend \$0.05 for each QA pair costing a total of \$$550$. The crowd-sourced cleaning was of high quality; on a set of $280$ expert annotated question-answer pairs, the crowd had an agreement score of 97\%. The resulting dataset is summarized in Table \ref{tab:QA_PAIRS}. 

\begin{table}[ht]
\scriptsize
\setlength{\tabcolsep}{2pt}
\begin{tabular}{c|c|c|c|c|c|c}

\textbf{}           & \multicolumn{1}{l|}{ \textbf{\#Ques.}} & \multicolumn{1}{l|}{\textbf{\begin{tabular}[c]{@{}c@{}}  QA \\pairs\end{tabular}}} & \multicolumn{1}{|c}{\textbf{\begin{tabular}[c]{@{}c@{}}Tokens\\ per ques.\end{tabular}}} & \multicolumn{1}{|c|}{\textbf{\begin{tabular}[c]{@{}c@{}}\#QA Pairs \\with Hotels\end{tabular}}} & \multicolumn{1}{c|}{\textbf{\begin{tabular}[c]{@{}c@{}}\#QA Pairs \\with Restr.\end{tabular}}} & \multicolumn{1}{c}{\textbf{\begin{tabular}[c]{@{}c@{}}\#QA Pairs \\with Attr.\end{tabular}}} \\ \hline \hline
 \textbf{Training}& 18,531                                  & 38,586 &  73.30  & 4,819 & 30,106 & 3,661 \\
 \textbf{Validation}& 2119                                 & 4,196                                      & 70.67 & 585 & 3267 & 335 \\
 \textbf{Test}& 2,173                                  & 4,342 & 70.97 & 558 & 3,418 & 366
\end{tabular}
\caption{QA Pairs in train,validation and test sets} \label{tab:QA_PAIRS}
\vspace{-5ex}
\end{table}

\subsection{Data Characteristics}


 In our dataset, the average number of tokens in each question is  approximately $73$, which is comparable to the document lengths for some existing QA tasks. Additionally, our entity documents are larger than the documents used in existing QA datasets (See Table \ref{tab:related}) -- they contain $3,266$ tokens on average. Lastly, answering any question requires studying all the possible entities in a given city -- the average number of candidate answer entities per question is more than $5,300$ which further highlights the challenges of scale for this task. 

\begin{table}[ht]
\scriptsize
\begin{tabular}{|c|c|l|} 
\hline
\textbf{Feature}                                                                                                                                                          & \textbf{\%}          & \multicolumn{1}{c|}{\textbf{Examples of Phrases in Questions}}                                                                                                                                                                                                                                                                                                                                                                                                                                                                                                                                                                                                                     \\ \hline \hline
\textbf{\begin{tabular}[c]{@{}c@{}}Budget \\ constraints\end{tabular}}                 & 23                   & \textit{\begin{tabular}[c]{@{}l@{}}good prices, \\ money is a bit of an issue \\ maximum of \$250 ish in total\end{tabular}}                                                                                                                                                                                                                                                                                                                                                                                                                                                               \\ \hline
\textbf{\begin{tabular}[c]{@{}c@{}}Temporal \\ elements\end{tabular}}                                          & 21                   & \textit{\begin{tabular}[c]{@{}l@{}}play ends around at 22:00 (it's so late!) \\.. dinner before the show,
\\ theatre for a Saturday night \\ open christmas eve 
\end{tabular}}                                                                                                                                                                                                                                                                                                                                             \\ \hline
\textbf{\begin{tabular}[c]{@{}c@{}}Location\\ constraint\end{tabular}}                 & 41                   & \textit{\begin{tabular}[c]{@{}l@{}}dinner near Queens Theatre, \\ staying in times square;would like it close,\\ 
options in close proximity (walking distant)\\ easy to get to from the airport 
\end{tabular}}                                                                                                                                                                                                \\ \hline
\textbf{\begin{tabular}[c]{@{}c@{}}Example entities \\ mentioned\end{tabular}}         & 8                   & \textit{\begin{tabular}[c]{@{}l@{}}found this one - Duke of Argyll \\ done the Wharf and Chinatown, 
\\ no problem with Super 8\end{tabular}}             \\ \hline
\textbf{\begin{tabular}[c]{@{}c@{}}Personal \\ preferences \end{tabular}} & 61                   & \textit{\begin{tabular}[c]{@{}l@{}}something unique and classy, \\ am not much of a shopper, \\ love upscale restaurants, \\ avoid the hotel restaurants, \\ Not worried about eating healthy 
\\ out with a girlfriend for a great getaway 
\end{tabular}} \\ \hline

\end{tabular} 
\caption{Classification of Questions - a qualitative study on 100 random samples. (\%) does not sum to 100; Questions may exhibit more than one feature. }
\label{tab:questions}
\vspace{-2ex}
\end{table}

Our dataset contains QA pairs for $50$ cities. The total number of entities in our dataset is $216,033$. In almost every city, the most common entity class is restaurants. On average, each question has $2$ gold answers extracted. 
61\% of the questions contain personal preferences of users, 23\% of the questions contain constraints on budgetary constraints, while 41\% contain locative constraints (Table \ref{tab:questions}). 
Details about the knowledge source are summarized in Table \ref{tab:knowledge}. Samples of review documents of entities, QA pairs as well as additional characteristics of the dataset are available for reference in the supplementary material. 

\begin{table}[ht]
\center
\scriptsize
\begin{tabular}{l|l}
Avg \# Tokens            & 3266 \\ \hline
Avg \# Reviews           & 69   \\ \hline
Avg \# Tokens per Review & 47   \\ \hline
Avg \# Sentences         & 263 
\end{tabular}
\vspace{-2ex}
\caption{Summarized statistics: Knowledge source consisting of $216,033$ entities and their reviews}
\label{tab:knowledge}
\vspace{-5ex}
\end{table}


\section{The Cluster-Select-Rerank Model} \label{sec:model}
\begin{figure*}[ht]
 {\footnotesize
 \hspace*{-0.7cm} 
   \includegraphics[scale=0.5]{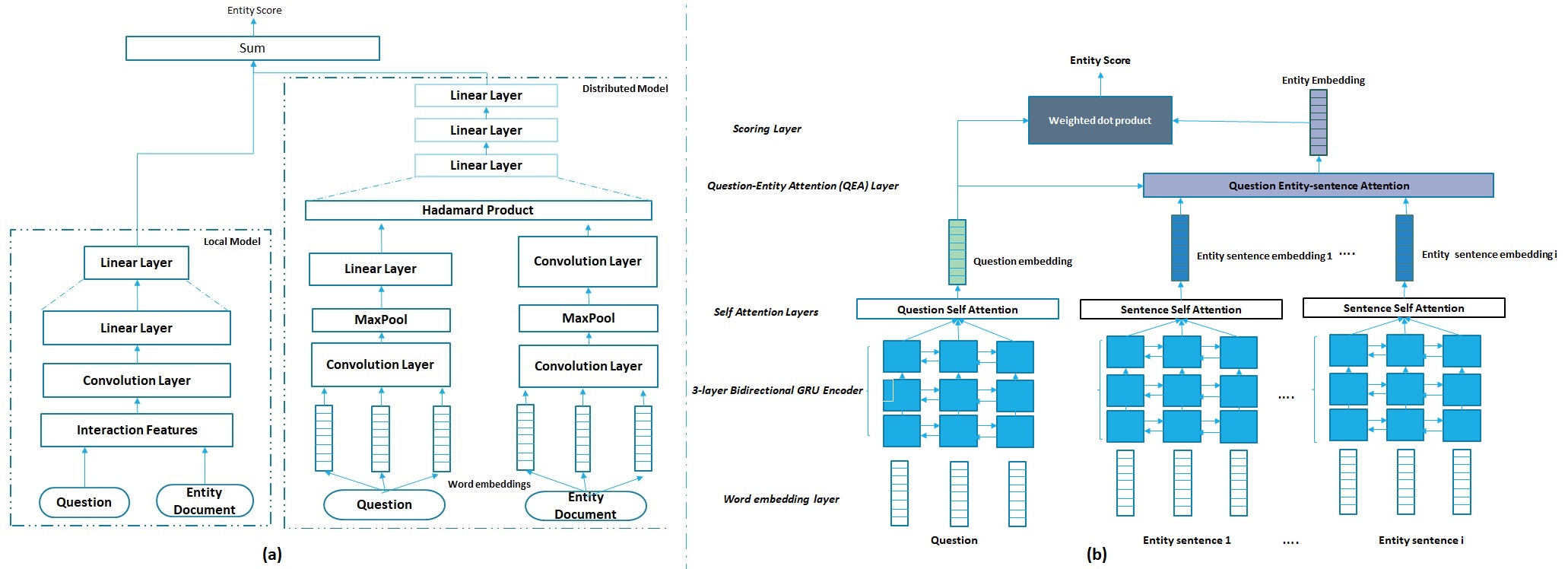}
   \vspace*{-4ex}
   \caption{(a) The Duet retrieval model \cite{DUET,DUET2} (b)Reasoning network used to re-rank candidates shortlisted by the Duet model. Units in the same colour indicate parameter sharing in (b).}
   \label{fig:siamese}
 
 }
\vspace*{-3ex}
 \end{figure*}
We now describe our model that trains on our dataset to answer a new question. 
Our model uses a {\em cluster-select-rerank} approach and combines benefits of both IR and QA architectures. We refer to it as $\csr$. 
It consists of three major components: (1) a {\bf clustering} module to generate representative entity documents, (2) a fast scalable retrieval model that {\bf selects} candidate answers and reduces the search space, and (3) a QA-style {\bf re-ranker} that reasons over the selected answers and scores them to return the final answer. We now describe each component in detail.


\subsection{\emph{Cluster:} ~Representative Entity Document Creation}
As stated previously, entity documents in our dataset are much larger than documents used by previous QA tasks. In order to make training a sufficiently expressive neural model tractable, 
\csr\ first constructs smaller representative documents\footnote{representative documents are a set of review sentences} for each entity using the full entity documents (containing all reviews for an entity). It encodes each review sentence using the pre-trained universal sentence encoder (USE) \cite{USE2018} to generate sentence embeddings. It then clusters sentences within each document and uses the top-$k$(nearest to the cluster centroid) sentences from each cluster to represent the entity. In our experiments we use $k=10$ and generate $10$ clusters per entity, thus reducing our document size to $100$ sentences each. This constitutes an approximately $70$\% reduction in document size. We note that despite this reduction our problem continues to be large-scale. This is because our documents are still larger than those used in most QA tasks and before returning an answer to a question, the system still has to explore over $500$ times\footnote{Most QA tasks with large answer spaces are able to filter (reduce to top-10) candidates using TFIDF-style methods.} more documents, as compared to most existing QA tasks.

\subsection{\emph{Select:} ~Shortlisting Candidate Answers} 
In this step, \csr\ trains a neural retrieval model with the question as the query and representative entity documents as the text corpus. As its retrieval model, it uses the recently improved Duet \cite{DUET2} network. Duet is an interaction-based neural network that compares elements of the question with different parts of a document and then aggregates evidence for relevance. It uses both local as well as distributed representations to capture lexical and semantic features. It is quite scalable for our task, since its neural design is primarily based on CNNs (Figure \ref{fig:siamese} (a)).

Duet is trained over the QA-pair training dataset and $10$ randomly sampled negative examples and uses cross-entropy loss. Duet can be seen as ranking the full candidate answer space for a given question, since it scores each representative entity document. \csr\ selects the top-30 candidate entities from this ranked list for a deeper reading and reasoning, as described in the next section.



\subsection{\emph{Rerank:} ~Answering over Selected Candidates}
In this step, our goal is to perform careful reading and reasoning over the shortlisted candidate answers to build the best QA system. 
The \csr\ implements a model for re-ranking based on Siamese network\cite{Siamese1,Siamese2} with recurrent encoding and attention-based matching. 

\noindent {\bf Input Layer:} It uses 128-dimensional word2vec embeddings \cite{word2vec} to encode each word of a question and a representative entity document. It uses a three layer bi-directional GRU \cite{GRU}, which is shared between the question and the review sentence encoder.

\noindent {\bf Self Attention Layer:} It learns shared self-attention (intra-attention) weights for questions and representative entity documents \cite{intra-attention} and generates attended embedding representations for both. 

\noindent {\bf Question-Entity Attention (QEA) Layer:} 
In order to generate an entity embedding, it attends over the sentence embeddings\footnote{obtained from the Self Attention Layer} of its representative entity document, with respect to the question \cite{Luong}. This helps identify ``important'' sentences and the sentence embeddings are then combined based on their attention weights to create the entity embedding. Thus, the entity embeddings are question-dependent.

\noindent{\bf Scoring Layer:} Finally, given a question and the entity embedding, the model uses a weighted dot product between the two vectors to generate the score that is used to compute the max-margin loss. The model is summarized in Figure \ref{fig:siamese} (b).

The network is trained by sampling $10$ negative (incorrect answer) entities for each question-answer pair and using hinge loss. We improve model training by employing curriculum learning and present harder negative samples returned by a simpler version of the ranker. For details and hyper-parameter settings, please refer to the Supplementary notes.


\section{Experiments}\label{sec:experiments}

We ask the following questions in our experiments: 
(1) What is quality of data collected by our automated method?
(2) What is the performance of the \csr\ model compared to other baselines for this task?
(3) How does the \csr\ model compare with neural IR and neural QA models?
(4) What are the characteristics of questions correctly answered by our system?

\subsection{Qualitative Study: Data} \label{sec:qual}
We studied $450$ QA pairs of the train-set\footnote{Note: this set is not cleaned by crowd-sourced workers}, representing approximately $1$\% of the dataset, for errors in the automated data collection process. We found that our high precision filtering rules have an answer extraction accuracy of $82$\%. The errors can be traced to one of four major causes (i) ({\bf $16$\%}) Entity name was a generic English word (e.g. ``The Park'') (ii) ({\bf $27$\%})  Entity matched another entity in the answer response which was not intended to be the answer entity to the original question. (e.g. Starbucks in "next to Starbucks") 
 (iii) ({\bf $31$\%}) Entity matched another entity with a similar name but of a different target class (e.g. hotel with same name instead of restaurant).
 (iv) ({\bf $13$\%}) Failing to detect negations/negative sentiment (e.g. an entity mention in a post where the user says ``{\em i wouldn't go there for the food}''. (v) The remaining {\bf $13$\%} of the errors were due to errors such as invalid questions (non-entity seeking), or incorrect answers provided by the forum users. 
 
We find that the extraction accuracy is comparable to that seen in some existing datasets such as TriviaQA \cite{TriviaQA}. However, as described previously we crowd-source and clean\footnote{97\% agreement with experts} both our test and validation sets to allow accurate assessment and bench-marking of model performance of any system designed for our task.

\subsection{Models for comparison}
We began by trying to adapt traditional reading comprehension QA models such as BiDAF \cite{BIDAF} for our task, but we found they were infeasible to run -- just $1$ epoch of training using $10$ negative samples per QA pair, and our representative entity documents, took BiDAF over $43$ hours to execute on $4$ K-$80$ GPUs. Running a trained BiDAF model on our test data would take even longer and was projected to require over $220$ hours. Similarly, we also tried using models based on BERT fine-tuning, but again, it did not scale for our task. In the absence of obvious scalable QA baselines, we 
compare the performance of \csr\  with other baselines for our task. 

\noindent{\bf Random Entity Baseline:} Returns a random ranking of the candidate answer space.

\noindent{\bf Ratings Baseline:} Returns a global (question-independent) ranking of candidate entities based on user review ratings of entities.

\noindent{\bf BM25 Retrieval:} We index each entity along with its reviews into Lucene\footnote{\emph{http://lucene.apache.org/}}. Each question is transformed into a query using the default query parser that removes stop words and creates a disjunctive term query. Entities are scored and ranked using BM25 ranking \cite{BM25}. Note that this baseline is considered a strong baseline for information retrieval (IR) and is, in general, better than most neural IR models \cite{BM25Good}.  

\noindent{\bf Review-AVG Model: }
This baseline uses averaged vector embeddings of the review sentences to represent each document - we use universal sentence embeddings (USE) \cite{USE2018} to pre-compute vector representations for each sentence and average them to create a document representation. Questions are encoded using a self-attended bi-directional GRU \cite{intra-attention} to generate a question representation. An entity is scored via a weighted dot product between question and document embeddings.
\subsubsection{Ablation Models}
\noindent{\bf \rsr:} This model highlights the value of the clustering step and the creation of representative entity documents.  We replace the clustering phase of our \csr model and use $100$ randomly-selected review-sentences to represent entities. We also tried to create a model that creates document representations by selecting $100$ sentences from an entity document by indexing them in Lucene and then using the question as a query. However, this method, understandably, returned very few sentences -- the questions (query) are longer than a sentence on average and the lexical gap is too big to overcome with simple expansion techniques.  Lastly, if we give the full entity document instead of a representative one, the neural select-rerank model cannot be trained due to GPU memory limitations.

\noindent{\bf \cs\ :} This model returns answers by running the neural information retrieval model, Duet, on the clustered representative documents. This model is effectively the \csr\ model but without re-ranking.

\noindent{\bf \crq\ :} This model returns answers by running the reasoner directly on the clustered representative documents. Thus, this model does not use neural IR to select and reduce the candidate search space.


\subsection{\bf Metrics for Model evaluation}
We use Accuracy@N metrics for evaluating a QA system. 
For a question $q$, let the set of top ranked $N$ entities returned by the system be $E_N$, and let the correct (gold) answer entities for the question be denoted by set $G$. We give credit to a system for Accuracy@N if the sets $E_N$ and $G$ have a non-zero intersection. We also use the standard mean reciprocal rank (MRR) metric. To compute MRR score we only consider the highest ranked gold answer (if multiple gold answers exist for a question).

\subsection{Results}

Table \ref{tab:expts} compares \csr\ against other models. We find that all non-neural baselines perform poorly on the task. Even the strong baseline of BM25 retrieval, which is commonly used in retrieval tasks, is not as effective for this dataset. Methods such as BM25 are primarily aimed at addressing challenges of semantic gap while in our task, answers require {\em reasoning} over subjective opinions in entity documents. 
We also observe that the performance of the neural model, Review-AVG, is comparable to that of BM25. 

\begin{table}[ht]
\centering
\footnotesize
\begin{tabular}{|l||r|r|r|r|}
\hline

\textbf{Method}                                              & \textbf{Acc@3} & \textbf{Acc@5} & \textbf{Acc@30 } & \textbf{MRR} \\ \hline \hline
\textbf{Random} & 0.32               & 0.58               &    3.78                             &   0.007            \\ 
\textbf{\begin{tabular}[c]{@{}c@{}}Ratings\\ \end{tabular}}  &         0.37        & 0.92                &   3.33                           & 0.007          \\  
\textbf{BM25 }   & 6.72                                           &  9.98               &  30.60              &  0.071                                              \\ 
\textbf{Review-AVG}                                                                 &  {7.87}              &  {11.83}              &            {30.65}     &     {0.084}         \\ 
\textbf{\rsr}   & 10.22          & 14.63         &     36.99        & 0.104     \\ 
\textbf{\crq}  &  {16.89}               &  {23.75}               &            {52.51}     &     {0.159} \\
\textbf{\cs} & {17.25}               &  {23.01}               &            {52.65}     &     {0.161}            \\  

\textbf{$\csr$}   & {\bf 21.44}            & {\bf 28.20}               &  {\bf 52.65}               & {\bf 0.186}       \\ \hline     


\end{tabular}
\caption{Performance of different systems including the $\csr$  model on our task. Accuracy reported in \%.} \label{tab:expts}
\vspace*{-2ex}
\end{table}

The $\rsr$ model that uses randomly sampled review-sentences, has a low Acc@3 of 10.22 \%. In contrast, both the $\cs$ and $\crq$ models, that use the clustered representative entity-documents have higher accuracy than $\rsr$. Our final model $\csr$, has an Acc@3 of approximately 21.44\% (last row).

We also find that \csr\ does better than \crq\ . We attribute the gain in using \csr\ over \crq\ to the fact 
that training the reasoner is compute intensive, and it is unable to see many hard negative samples for a question even after a long time of training. Due to this it optimizes its loss on the negatives seen during training, but may not perform well when the full candidate set is provided. On the other hand, in the complete \csr\ model, the {\em select} module shortlists good candidates apriori and the reasoner's job is limited to finding the best ones from the small set of good candidates.  

Comparing  \csr\ and \cs\ suggests that, while the scalable matching of Duet is useful enough for filtering candidates, it is not good enough to return the best answer. On the other hand the \csr\ model has a reasoner specifically trained to re-rank a harder set of filtered candidates and hence performs better. 

Overall, we find that each component of \csr\ is critical in its contributing towards its performance on the task. Moreover, strong IR only ($\cs$) and QA only baselines (\crq) are not as effective as their combination in \csr. 

\subsection{Answering Characteristics}
\begin{table}[ht]
\scriptsize
\center
\begin{tabular}{|c|c|c|c|c|}
\hline
   \textbf{\begin{tabular}[c]{@{}c@{}}Candidate Space\\Size \end{tabular}}  & \textbf{\begin{tabular}[c]{@{}c@{}}No. of \\ Questions \end{tabular}}  & \bf \cs\ & \bf \crq  & \bf \csr \\
 \hline \hline
 \textless{}=1000 & 631   & 28.69 & 30.27 & {\bf 32.49} \\
 \textgreater{}1000  & 1542 & 12.58 & 11.41 & {\bf 16.93} \\
\hline
\end{tabular}
\caption{Test set performance (Acc@3 in \%) of ablation systems on questions with different candidate answer space sizes.}
\label{tab:spaces}
\vspace{-5ex}
\end{table}

Table \ref{tab:spaces} breaks down the performance of systems based on size of the candidate space encountered while answering. In questions where the candidate space is relatively smaller (<1000), we find $\crq$ model has slightly better performance than the $\cs$ model. However, in large candidate spaces we find the $\cs$ model is more effective in pruning the candidate search space and performs better than the $\crq$ model. The $\csr$ model outperforms both systems regardless of candidate space size, highlighting the benefit of our method.

\subsection{Qualitative Study: Answering System}
Since we rely on automated methods used to construct the dataset, it is likely that our precision-oriented rules for data-set creation erroneously exclude some entity-answers originally recommended by forum users. In addition, there may also be alternative recommendations that were not part of the original forum responses, but may be valid alternatives. Therefore, we assess whether metrics computed using the gold-entity answers as reference answers, correlate with human relevance judgements on the top-3 answers returned by a system.

\begin{table}[]

\scriptsize
\center
\begin{tabular}{|l||c|c|}
\hline                                      & \multicolumn{1}{|l|}{\textbf{Human Scores}} & \multicolumn{1}{|l|}{\textbf{Machine Scores}} \\ \hline
\multicolumn{1}{|l|}{\textbf{Method}} & \multicolumn{1}{c|}{\textbf{Acc@3}}       & \textbf{Acc@3}                              \\ \hline
\multicolumn{1}{|l|}{\textbf{CR}}     & \multicolumn{1}{c|}{50.0}                 & 19.79                                       \\
\multicolumn{1}{|l|}{\textbf{CS}}     & \multicolumn{1}{c|}{63.51}                & 22.92                                       \\
\multicolumn{1}{|l|}{\textbf{CSR}}    & \multicolumn{1}{c|}{\textbf{65.63}}       & \textbf{33.33}                              \\ \hline
\end{tabular}
\caption{Performance of different systems including the $\csr$  model on our task as measured using human judgements (Human Scores) and gold-reference data (Machine Scores). Accuracy reported in \%.} \label{tab:human}
\end{table}
We randomly select $100$ questions from the validation data and use the top-3 answers returned from three models, $\cs$, $\crq$ and $\csr$ for a qualitative study. The human evaluators part of this study are blind to the models returning the answers and we present each question-recommendation pair independently and in random order. We ask the evaluators to manually query a web-search engine and asses if each question-recommendation pair (returned by a model) adequately matches the requirements of the user posting that question. Specifically, the  evaluators are asked to rank an answer correctly, if in their judgement, a candidate answer would have been one that they would recommend to a user based on the information they find on the web. To keep the real-world nature of the task intact we do not ask them to refer to specific websites or pages but suggest that they consider reviews, ratings, location, popularity, budget, convenience of access/transportation, timings when marking a candidate answer as a valid recommendation. Thus, evaluating whether an entity-answer returned is correct is subjective and time consuming. Based on these guidelines, two evaluators assessed a different set of $100$ unseen questions-recommendation pairs with over $1300+$ entities and we found the inter-annotator agreement on relevance judgements to be $0.79$ . 

\subsection{Results}

The results of the qualitative evaluation on the $300$ QA Pairs ($100$ questions) from the validation data are summarized in Table \ref{tab:human}. As can be seen from the table, the absolute performance of the systems as measured by the human annotators is higher indicating the presence of false negatives in the dataset. 
In order to assess whether performance improvements measured using our gold-data correlate with human judgements on this task, we compute the Spearman's rank coefficient\footnote{https://en.wikipedia.org/wiki/Spearman\%27s\_rank\\\_correlation\_coefficient} between the human assigned $Acc@N$ scores and machine-evaluated $Acc@N$ scores. We compute pair-wise correlation coefficients between $\csr$, $\cs$ and $\crq$, and we find there is moderately positive correlation \cite{correlation} with high confidence between the human judgements and gold-data based measurements for both Acc@3 ($\bar{\rho}=0.39$, p-value<0.0009) as well as on Acc@5 ($\bar{\rho}=0.32$ p-value<0.04). Please see appendix for more details.

\subsection{Error Analysis}

We conducted an error analysis of the \csr\ model using the results of the human evaluation. We found that nearly $35$\% of the errors made were on questions involve location constraints while, $9$\% of the errors were due to either budgetary or temporal constraints not being satisfied. 
(See Appendix for more details)

\section{Conclusion}
In the spirit of defining a question answering challenge that is closer to a real-world QA setting,  we introduce the novel task of identifying the correct entity answer to a given user question based on a collection of unstructured reviews describing entities. We harvest a dataset of over 47,000 QA pairs, which enables end to end training of models.  

The biggest challenge in this dataset is that of scalability. Our task requires processing $500$ times more documents per question than most existing QA tasks, and individual documents are also much larger in size.
In response, we develop a cluster-select-rerank architecture that brings together neural IR and QA models for an overall good performance. 
Our best system registers a 25\% relative improvement over our baseline models. However, a correct answer is in top-3 for only 21\% of the questions, which points to the difficulty of the task.

We believe that further research on this task will significantly improve the state-of-the-art in question answering. Neuro-symbolic methods that reason on locative and budgetary constraints could be an interesting direction of future work. These types of questions constitute nearly 64\% of the user constraints specified in questions in our dataset. 
We will make resources from this paper available for further research (please see appendix for more details).

\section{Acknowledgements}
 We would like to thank Yatin Nandwani, Sumit Bhatia, Dhiraj Madan, Dinesh Raghu, Sachindra Joshi, Shashank Goel, Gaurav Pandey, Dinesh Khandelwal for their helpful suggestions during the course of this work. We would like to thank Shashank Goel for re-implementing the data collection scripts and maintaining the GitHub repository. We would also like to acknowledge the IBM Research India PhD program that enables the first author to pursue the PhD at IIT Delhi. This work is supported by an IBM AI Horizons Network grant, IBM SUR awards,  Visvesvaraya faculty awards by Govt. of India to both Mausam and Parag as well as grants by Google,  Bloomberg  and 1MG  to Mausam.

\bibliography{arxiv-main.bib}

\begin{thebibliography}{35}
\expandafter\ifx\csname natexlab\endcsname\relax\def\natexlab#1{#1}\fi

\bibitem[{Agichtein et~al.(2015)Agichtein, Carmel, Pelleg, Pinter, and
  Harman}]{LiveQA}
Eugene Agichtein, David Carmel, Dan Pelleg, Yuval Pinter, and Donna Harman.
  2015.
\newblock \href {http://trec.nist.gov/pubs/trec24/papers/Overview-QA.pdf}
  {Overview of the {TREC} 2015 liveqa track}.
\newblock In \emph{Proceedings of The Twenty-Fourth Text REtrieval Conference,
  {TREC} 2015, Gaithersburg, Maryland, USA, November 17-20, 2015}.

\bibitem[{Akoglu(2018)}]{correlation}
Haldun Akoglu. 2018.
\newblock User's guide to correlation coefficients.
\newblock \emph{Turkish Journal of Emergency Medicine}, 18:91 -- 93.

\bibitem[{Castelli et~al.(2019)Castelli, Chakravarti, Dana, Ferritto, Florian,
  Franz, Garg, Khandelwal, McCarley, McCawley, Nasr, Pan, Pendus, Pitrelli,
  Pujar, Roukos, Sakrajda, Sil, Uceda{-}Sosa, Ward, and Zhang}]{TechQA}
Vittorio Castelli, Rishav Chakravarti, Saswati Dana, Anthony Ferritto, Radu
  Florian, Martin Franz, Dinesh Garg, Dinesh Khandelwal, J.~Scott McCarley,
  Mike McCawley, Mohamed Nasr, Lin Pan, Cezar Pendus, John~F. Pitrelli, Saurabh
  Pujar, Salim Roukos, Andrzej Sakrajda, Avirup Sil, Rosario Uceda{-}Sosa, Todd
  Ward, and Rong Zhang. 2019.
\newblock \href {http://arxiv.org/abs/1911.02984} {The techqa dataset}.
\newblock \emph{CoRR}, abs/1911.02984.

\bibitem[{Cer et~al.(2018)Cer, Yang, Kong, Hua, Limtiaco, John, Constant,
  Guajardo{-}Cespedes, Yuan, Tar, Sung, Strope, and Kurzweil}]{USE2018}
Daniel Cer, Yinfei Yang, Sheng{-}yi Kong, Nan Hua, Nicole Limtiaco, Rhomni~St.
  John, Noah Constant, Mario Guajardo{-}Cespedes, Steve Yuan, Chris Tar,
  Yun{-}Hsuan Sung, Brian Strope, and Ray Kurzweil. 2018.
\newblock \href {http://arxiv.org/abs/1803.11175} {Universal sentence encoder}.
\newblock \emph{CoRR}, abs/1803.11175.

\bibitem[{Chen et~al.(2017)Chen, Fisch, Weston, and Bordes}]{drqa}
Danqi Chen, Adam Fisch, Jason Weston, and Antoine Bordes. 2017.
\newblock Reading {Wikipedia} to answer open-domain questions.
\newblock In \emph{Association for Computational Linguistics (ACL)}.

\bibitem[{Cheng et~al.(2016)Cheng, Dong, and Lapata}]{intra-attention}
Jianpeng Cheng, Li~Dong, and Mirella Lapata. 2016.
\newblock \href {https://doi.org/10.18653/v1/D16-1053} {Long short-term
  memory-networks for machine reading}.
\newblock In \emph{Proceedings of the 2016 Conference on Empirical Methods in
  Natural Language Processing}, pages 551--561, Austin, Texas. Association for
  Computational Linguistics.

\bibitem[{Cho et~al.(2014)Cho, van Merrienboer, Gulcehre, Bahdanau, Bougares,
  Schwenk, and Bengio}]{GRU}
Kyunghyun Cho, Bart van Merrienboer, Caglar Gulcehre, Dzmitry Bahdanau, Fethi
  Bougares, Holger Schwenk, and Yoshua Bengio. 2014.
\newblock \href {https://doi.org/10.3115/v1/D14-1179} {Learning phrase
  representations using {RNN} encoder{--}decoder for statistical machine
  translation}.
\newblock In \emph{Proceedings of the 2014 Conference on Empirical Methods in
  Natural Language Processing ({EMNLP})}, pages 1724--1734, Doha, Qatar.
  Association for Computational Linguistics.

\bibitem[{Choi et~al.(2018)Choi, He, Iyyer, Yatskar, Yih, Choi, Liang, and
  Zettlemoyer}]{QuAC}
Eunsol Choi, He~He, Mohit Iyyer, Mark Yatskar, Wen{-}tau Yih, Yejin Choi, Percy
  Liang, and Luke Zettlemoyer. 2018.
\newblock \href {https://aclanthology.info/papers/D18-1241/d18-1241} {Quac:
  Question answering in context}.
\newblock In \emph{Proceedings of the 2018 Conference on Empirical Methods in
  Natural Language Processing, Brussels, Belgium, October 31 - November 4,
  2018}, pages 2174--2184.

\bibitem[{Contractor et~al.(2020)Contractor, Patra, Mausam, and Singla}]{MSEQ2}
Danish Contractor, Barun Patra, Mausam, and Parag Singla. 2020.
\newblock \href {https://doi.org/10.1017/S1351324920000017} {Constrained bert
  bilstm crf for understanding multi-sentence entity-seeking questions}.
\newblock \emph{Natural Language Engineering}, page 1–23.

\bibitem[{Devlin et~al.(2019)Devlin, Chang, Lee, and Toutanova}]{BERT}
Jacob Devlin, Ming{-}Wei Chang, Kenton Lee, and Kristina Toutanova. 2019.
\newblock \href {https://aclweb.org/anthology/papers/N/N19/N19-1423/} {{BERT:}
  pre-training of deep bidirectional transformers for language understanding}.
\newblock In \emph{Proceedings of the 2019 Conference of the North American
  Chapter of the Association for Computational Linguistics: Human Language
  Technologies, {NAACL-HLT} 2019, Minneapolis, MN, USA, June 2-7, 2019, Volume
  1 (Long and Short Papers)}, pages 4171--4186.

\bibitem[{Dunn et~al.(2017)Dunn, Sagun, Higgins, G{\"{u}}ney, Cirik, and
  Cho}]{SearchQA}
Matthew Dunn, Levent Sagun, Mike Higgins, V.~Ugur G{\"{u}}ney, Volkan Cirik,
  and Kyunghyun Cho. 2017.
\newblock \href {http://arxiv.org/abs/1704.05179} {Searchqa: {A} new q{\&}a
  dataset augmented with context from a search engine}.
\newblock \emph{CoRR}, abs/1704.05179.

\bibitem[{Fan et~al.(2019)Fan, Jernite, Perez, Grangier, Weston, and
  Auli}]{LongFormQA}
Angela Fan, Yacine Jernite, Ethan Perez, David Grangier, Jason Weston, and
  Michael Auli. 2019.
\newblock \href {https://www.aclweb.org/anthology/P19-1346/} {{ELI5:} long form
  question answering}.
\newblock In \emph{Proceedings of the 57th Conference of the Association for
  Computational Linguistics, {ACL} 2019, Florence, Italy, July 28- August 2,
  2019, Volume 1: Long Papers}, pages 3558--3567.

\bibitem[{Hewlett et~al.(2016)Hewlett, Lacoste, Jones, Polosukhin, Fandrianto,
  Han, Kelcey, and Berthelot}]{WikiReading}
Daniel Hewlett, Alexandre Lacoste, Llion Jones, Illia Polosukhin, Andrew
  Fandrianto, Jay Han, Matthew Kelcey, and David Berthelot. 2016.
\newblock \href {https://www.aclweb.org/anthology/P16-1145/} {Wikireading: {A}
  novel large-scale language understanding task over wikipedia}.
\newblock In \emph{Proceedings of the 54th Annual Meeting of the Association
  for Computational Linguistics, {ACL} 2016, August 7-12, 2016, Berlin,
  Germany, Volume 1: Long Papers}.

\bibitem[{Joshi et~al.(2017)Joshi, Choi, Weld, and Zettlemoyer}]{TriviaQA}
Mandar Joshi, Eunsol Choi, Daniel~S. Weld, and Luke Zettlemoyer. 2017.
\newblock \href {https://doi.org/10.18653/v1/P17-1147} {Triviaqa: {A} large
  scale distantly supervised challenge dataset for reading comprehension}.
\newblock In \emph{Proceedings of the 55th Annual Meeting of the Association
  for Computational Linguistics, {ACL} 2017, Vancouver, Canada, July 30 -
  August 4, Volume 1: Long Papers}, pages 1601--1611.

\bibitem[{Kwiatkowski et~al.(2019)Kwiatkowski, Palomaki, Redfield, Collins,
  Parikh, Alberti, Epstein, Polosukhin, Kelcey, Devlin, Lee, Toutanova, Jones,
  Chang, Dai, Uszkoreit, Le, and Petrov}]{NaturalQuestions}
Tom Kwiatkowski, Jennimaria Palomaki, Olivia Redfield, Michael Collins, Ankur
  Parikh, Chris Alberti, Danielle Epstein, Illia Polosukhin, Matthew Kelcey,
  Jacob Devlin, Kenton Lee, Kristina~N. Toutanova, Llion Jones, Ming-Wei Chang,
  Andrew Dai, Jakob Uszkoreit, Quoc Le, and Slav Petrov. 2019.
\newblock \href
  {https://tomkwiat.users.x20web.corp.google.com/papers/natural-questions/main-1455-kwiatkowski.pdf}
  {Natural questions: a benchmark for question answering research}.
\newblock \emph{Transactions of the Association of Computational Linguistics}.

\bibitem[{Lai et~al.(2017)Lai, Xie, Liu, Yang, and Hovy}]{RACE}
Guokun Lai, Qizhe Xie, Hanxiao Liu, Yiming Yang, and Eduard Hovy. 2017.
\newblock \href {https://doi.org/10.18653/v1/D17-1082} {Race: Large-scale
  reading comprehension dataset from examinations}.
\newblock In \emph{Proceedings of the 2017 Conference on Empirical Methods in
  Natural Language Processing}, pages 785--794. Association for Computational
  Linguistics.

\bibitem[{Lai et~al.(2018)Lai, Bui, and Li}]{Siamese2}
Tuan~Manh Lai, Trung Bui, and Sheng Li. 2018.
\newblock \href {https://aclanthology.info/papers/C18-1181/c18-1181} {A review
  on deep learning techniques applied to answer selection}.
\newblock In \emph{Proceedings of the 27th International Conference on
  Computational Linguistics, {COLING} 2018, Santa Fe, New Mexico, USA, August
  20-26, 2018}, pages 2132--2144.

\bibitem[{Le and Mikolov(2014)}]{doc2vec}
Quoc Le and Tomas Mikolov. 2014.
\newblock \href {http://dl.acm.org/citation.cfm?id=3044805.3045025}
  {Distributed representations of sentences and documents}.
\newblock In \emph{Proceedings of the 31st International Conference on
  International Conference on Machine Learning - Volume 32}, ICML'14, pages
  II--1188--II--1196. JMLR.org.

\bibitem[{Louizos et~al.(2018)Louizos, Welling, and Kingma}]{Adam}
Christos Louizos, Max Welling, and Diederik~P. Kingma. 2018.
\newblock \href {https://openreview.net/forum?id=H1Y8hhg0b} {Learning sparse
  neural networks through l{\_}0 regularization}.
\newblock In \emph{6th International Conference on Learning Representations,
  {ICLR} 2018, Vancouver, BC, Canada, April 30 - May 3, 2018, Conference Track
  Proceedings}.

\bibitem[{Luong et~al.(2015)Luong, Pham, and Manning}]{Luong}
Thang Luong, Hieu Pham, and Christopher~D. Manning. 2015.
\newblock \href {http://aclweb.org/anthology/D/D15/D15-1166.pdf} {Effective
  approaches to attention-based neural machine translation}.
\newblock In \emph{Proceedings of the 2015 Conference on Empirical Methods in
  Natural Language Processing, {EMNLP} 2015, Lisbon, Portugal, September 17-21,
  2015}, pages 1412--1421.

\bibitem[{McDonald et~al.(2018)McDonald, Brokos, and
  Androutsopoulos}]{BM25Good}
Ryan McDonald, George Brokos, and Ion Androutsopoulos. 2018.
\newblock \href {https://aclanthology.info/papers/D18-1211/d18-1211} {Deep
  relevance ranking using enhanced document-query interactions}.
\newblock In \emph{Proceedings of the 2018 Conference on Empirical Methods in
  Natural Language Processing, Brussels, Belgium, October 31 - November 4,
  2018}, pages 1849--1860.

\bibitem[{Mihaylov et~al.(2018)Mihaylov, Clark, Khot, and
  Sabharwal}]{OpenBookQA2018}
Todor Mihaylov, Peter Clark, Tushar Khot, and Ashish Sabharwal. 2018.
\newblock Can a suit of armor conduct electricity? a new dataset for open book
  question answering.
\newblock In \emph{EMNLP}.

\bibitem[{Mikolov et~al.(2013)Mikolov, Sutskever, Chen, Corrado, and
  Dean}]{word2vec}
Tomas Mikolov, Ilya Sutskever, Kai Chen, Greg Corrado, and Jeffrey Dean. 2013.
\newblock \href {http://dl.acm.org/citation.cfm?id=2999792.2999959}
  {Distributed representations of words and phrases and their
  compositionality}.
\newblock In \emph{Proceedings of the 26th International Conference on Neural
  Information Processing Systems - Volume 2}, NIPS'13, pages 3111--3119, USA.
  Curran Associates Inc.

\bibitem[{Mitra and Craswell(2018)}]{IRSurvey}
Bhaskar Mitra and Nick Craswell. 2018.
\newblock \href
  {https://www.microsoft.com/en-us/research/publication/introduction-neural-information-retrieval/}
  {An introduction to neural information retrieval}.
\newblock \emph{Foundations and Trends® in Information Retrieval},
  13(1):1--126.

\bibitem[{Mitra and Craswell(2019)}]{DUET2}
Bhaskar Mitra and Nick Craswell. 2019.
\newblock An updated duet model for passage re-ranking.
\newblock \emph{arXiv preprint arXiv:1903.07666}.

\bibitem[{Mitra et~al.(2017)Mitra, Diaz, and Craswell}]{DUET}
Bhaskar Mitra, Fernando Diaz, and Nick Craswell. 2017.
\newblock Learning to match using local and distributed representations of text
  for web search.
\newblock In \emph{Proceedings of the 26th International Conference on World
  Wide Web}, pages 1291--1299. International World Wide Web Conferences
  Steering Committee.

\bibitem[{Nguyen et~al.(2016)Nguyen, Rosenberg, Song, Gao, Tiwary, Majumder,
  and Deng}]{MSMARCO}
Tri Nguyen, Mir Rosenberg, Xia Song, Jianfeng Gao, Saurabh Tiwary, Rangan
  Majumder, and Li~Deng. 2016.
\newblock {MS} {MARCO:} {A} human generated machine reading comprehension
  dataset.
\newblock In \emph{CoCo@NIPS}, volume 1773 of \emph{{CEUR} Workshop
  Proceedings}. CEUR-WS.org.

\bibitem[{Rajpurkar et~al.(2018)Rajpurkar, Jia, and Liang}]{SqUAD}
Pranav Rajpurkar, Robin Jia, and Percy Liang. 2018.
\newblock \href {https://aclanthology.info/papers/P18-2124/p18-2124} {Know what
  you don't know: Unanswerable questions for squad}.
\newblock In \emph{Proceedings of the 56th Annual Meeting of the Association
  for Computational Linguistics, {ACL} 2018, Melbourne, Australia, July 15-20,
  2018, Volume 2: Short Papers}, pages 784--789.

\bibitem[{Rao et~al.(2016)Rao, He, and Lin}]{Siamese1}
Jinfeng Rao, Hua He, and Jimmy Lin. 2016.
\newblock \href {https://doi.org/10.1145/2983323.2983872} {Noise-contrastive
  estimation for answer selection with deep neural networks}.
\newblock In \emph{Proceedings of the 25th ACM International on Conference on
  Information and Knowledge Management}, CIKM '16, pages 1913--1916, New York,
  NY, USA. ACM.

\bibitem[{Reddy et~al.(2018)Reddy, Chen, and Manning}]{CoQA}
Siva Reddy, Danqi Chen, and Christopher~D. Manning. 2018.
\newblock \href {http://arxiv.org/abs/1808.07042} {Coqa: {A} conversational
  question answering challenge}.
\newblock \emph{CoRR}, abs/1808.07042.

\bibitem[{Robertson and Zaragoza(2009)}]{BM25}
Stephen Robertson and Hugo Zaragoza. 2009.
\newblock \href {https://doi.org/10.1561/1500000019} {The probabilistic
  relevance framework: Bm25 and beyond}.
\newblock \emph{Found. Trends Inf. Retr.}, 3(4):333--389.

\bibitem[{Seo et~al.(2016)Seo, Kembhavi, Farhadi, and Hajishirzi}]{BIDAF}
Min~Joon Seo, Aniruddha Kembhavi, Ali Farhadi, and Hannaneh Hajishirzi. 2016.
\newblock \href {http://arxiv.org/abs/1611.01603} {Bidirectional attention flow
  for machine comprehension}.
\newblock \emph{CoRR}, abs/1611.01603.

\bibitem[{Trischler et~al.(2016)Trischler, Wang, Yuan, Harris, Sordoni,
  Bachman, and Suleman}]{NewsQA}
Adam Trischler, Tong Wang, Xingdi Yuan, Justin Harris, Alessandro Sordoni,
  Philip Bachman, and Kaheer Suleman. 2016.
\newblock \href {http://arxiv.org/abs/1611.09830} {Newsqa: {A} machine
  comprehension dataset}.
\newblock \emph{CoRR}, abs/1611.09830.

\bibitem[{Yang et~al.(2018)Yang, Qi, Zhang, Bengio, Cohen, Salakhutdinov, and
  Manning}]{HotpotQA}
Zhilin Yang, Peng Qi, Saizheng Zhang, Yoshua Bengio, William~W. Cohen, Ruslan
  Salakhutdinov, and Christopher~D. Manning. 2018.
\newblock {HotpotQA}: A dataset for diverse, explainable multi-hop question
  answering.
\newblock In \emph{Proceedings of the Conference on Empirical Methods in
  Natural Language Processing (EMNLP)}.

\bibitem[{Zhang et~al.(2018)Zhang, Wu, He, Liu, and Su}]{MedQA}
Xiao Zhang, Ji~Wu, Zhiyang He, Xien Liu, and Ying Su. 2018.
\newblock \href
  {https://www.aaai.org/ocs/index.php/AAAI/AAAI18/paper/view/16582} {Medical
  exam question answering with large-scale reading comprehension}.
\newblock In \emph{Proceedings of the Thirty-Second {AAAI} Conference on
  Artificial Intelligence, (AAAI-18), the 30th innovative Applications of
  Artificial Intelligence (IAAI-18), and the 8th {AAAI} Symposium on
  Educational Advances in Artificial Intelligence (EAAI-18), New Orleans,
  Louisiana, USA, February 2-7, 2018}, pages 5706--5713.

\end{thebibliography}
\bibliographystyle{acl_natbib}
\appendix
\setcounter{section}{7}
\section{Appendix}
The supplementary information contains the following:
\begin{itemize}
    \item Section \ref{sec:amt} describes details about the crowd-sourcing task used to clean the validation and test sets.
    \item Section \ref{sec:qual-question-study} presents a qualitative study characterizing the questions seen in our dataset.
    \item Section \ref{sec:hyper} describes the hyper-parameter settings used in our models.
    \item Section \ref{sec:answer-characteristics} describes how the size of candidate search space for each entity class affects answering accuracy of $\cs$, $\crq$ and $\csr$ systems.
    \item Section \ref{sec:re-rank} studies the effect of the number of candidates to be re-ranked affects the performance of the $\csr$ models.
    \item Section \ref{sec:curr} describes experiments indicating how generating better samples for training with curriculum learning affects the performance of the re-ranker ($\crq$).
    
    \item Section \ref{supp:corr} presents details of the correlation study on human-assigned relevance judgements and scores computed by using the automatically extracted data as gold-data.
    
    \item Section \ref{sec:answer-error} presents an error analysis performed on the answers returned by the $\csr$ system. 
    
    \item Section \ref{sec:datagen} describes the scripts being released as part of this work for generating the dataset as well scripts that allow users to collect QA pairs for new cities.
    
    \item Section \ref{sec:data-stats} contains some detailed statistics of the train, test and validation sets including the city and answer-entity class wise distributions.
\end{itemize}

 \subsection{Crowd-sourcing task} \label{sec:amt}
\begin{figure*}[ht]

 {\footnotesize
 \hspace{-0.5cm}
   \includegraphics[scale=0.5]{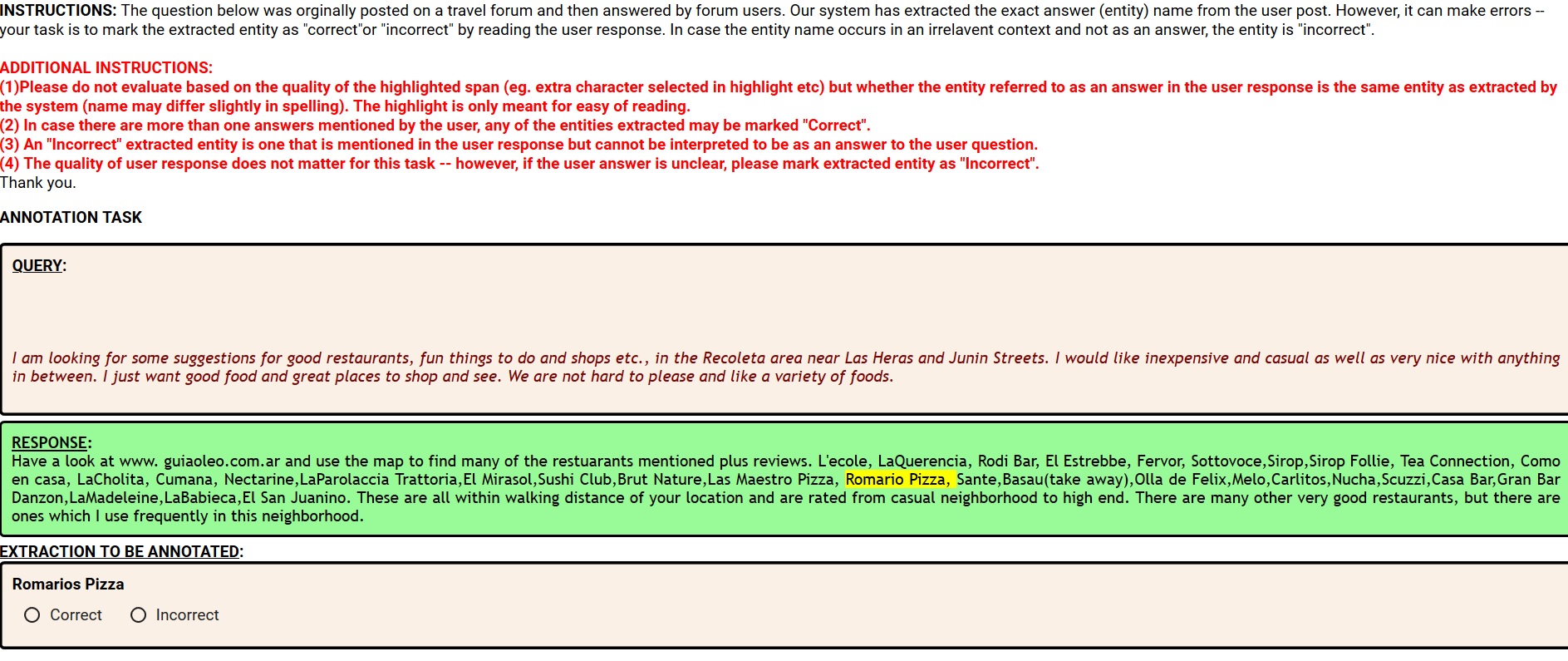}
   \caption{Human Intelligence Task (HIT) set up on Amazon Mechanical Turk to clean test and validation sets.}
   \label{fig:crowd}
 
 }
 \end{figure*}

In order to generate a clean test and validation set for accurate benchmarking we ask crowd-sourced workers. We use the Amazon Mechanical Turk(AMT)\footnote{http://requester.mturk.com} for crowd-sourcing. Workers are presented with a QA-pair, which includes the original question, an answer-entity extracted by our rules and the original forum post response thread where the answer entity was mentioned. Workers are then asked to check if the extracted answer entity was mentioned in the forum responses as an answer. Figure \ref{fig:crowd} shows an example of the task set up on AMT.

We spend \$0.05 for each QA pair costing a total of $550$. The crowd-sourced cleaning was of high quality -- on a set of $280$ expert annotated question-answer pairs, the crowd had an agreement score of 97\%. 

\subsection{Qualitative Study: Questions } \label{sec:qual-question-study}
\begin{table*}[ht]
\scriptsize
\begin{tabular}{|c|c|c|l|} 
\hline
\textbf{Feature}                                                                               & \textbf{Definition}                                                                                               & \textbf{\%}          & \multicolumn{1}{c|}{\textbf{Example of phrases}}                                                                                                                                                                                                                                                                                                                                                                                                                                                                                                                                                                                                                     \\ \hline \hline
\textbf{\begin{tabular}[c]{@{}c@{}}Budget \\ constraints\end{tabular}}                 & \begin{tabular}[c]{@{}c@{}}Explicit or implicit \\ mention of budgetary \\ constraints\end{tabular}               & 23                   & \textit{\begin{tabular}[c]{@{}l@{}}good prices, money is a bit of an issue, maximum of \$250 ish in total,\\ price isn't really an issue but want good value, not too concerned on price\end{tabular}}                                                                                                                                                                                                                                                                                                                                                                                                                                                               \\ \hline
\textbf{\begin{tabular}[c]{@{}c@{}}Temporal \\ elements\end{tabular}}                  & \begin{tabular}[c]{@{}c@{}}Time-sensitive \\ requirements/constraints\end{tabular}                                & 21                   & \textit{\begin{tabular}[c]{@{}l@{}}play ends around at 22:00 (it's so late!) ......dinner before the show, \\ pub from around 6-8.30, theatre for a Saturday night,suggestions for new year, \\ dinner on Friday night,  mid-day Sunday dinner (on Easter), talking almost midnight, \\ open christmas eve\end{tabular}}                                                                                                                                                                                                                                                                                                                                             \\ \hline
\textbf{\begin{tabular}[c]{@{}c@{}}Location\\ constraint\end{tabular}}                 & \begin{tabular}[c]{@{}c@{}}Entities need to \\ fulfill geographical constraints\end{tabular}                      & 41                   & \textit{\begin{tabular}[c]{@{}l@{}}dinner near Queens Theatre, staying in times square - so would like somewhere close by,\\ suggest somewhere near \textless{}LOC\textgreater{}, ..\textless{}LOC\textgreater{}.. restaurants in this area, \\ options within close proximity (walking distant), easy to get to from the airport, \\ Penn Quarter area, downtown restaurant, not too far from our hotel, \\ dont mind going to other areas of the city\end{tabular}}                                                                                                                                                                                                \\ \hline
\textbf{\begin{tabular}[c]{@{}c@{}}Example entities \\ mentioned\end{tabular}}         & \begin{tabular}[c]{@{}c@{}}Entities a user mentions\\ as examples of what they want\\  or dont want.\end{tabular} & 8                   & \textit{\begin{tabular}[c]{@{}l@{}}found this one - Duke of Argyll, done the Wharf and Chinatown, \\ Someone suggested Carmine's but they are totally booked, avoiding McDonalds,\\ no problem with Super 8,\end{tabular}}                                                                                                                                                                                                                                                                                                                                                                                                                                           \\ \hline
\textbf{\begin{tabular}[c]{@{}c@{}}Personal \\ preferences / constraints\end{tabular}} & User specific constraints                                                                                         & 61                   & \textit{\begin{tabular}[c]{@{}l@{}}something unique and classy, am not much of a shopper, love upscale restaurants, \\ avoid the hotel restaurants, Not worried about eating healthy, best seafood pasta, \\ large portion, traditional American-style breakfast, some live music, \\ stay away from the more touristy sort of place, go to dinner and dress up, \\ preferably not steak and not ultra-expensive, dont mind if its posh and upmarket,  \\ quick bite and a drink or two, ethnic options, vegetarian diet, 7 adults 20s-40s,\\  out with a girlfriend for a great getaway, american or italian cuisine, \\ nice restaurant to spoil her, places to do some surfing \end{tabular}} \\ \hline

\end{tabular} 
\caption{Classification of Questions. (\%) does not sum to 100, because questions may exhibit more than one feature.}
\label{tab:questionssupp}
\end{table*}


We analyzed $100$ questions and summarize their characteristics in Table \ref{tab:questionssupp}. As expected, most questions (61\%) have user-specific preferential constraints that govern the characteristics of the answer entity to be returned. As can be seen in phrases extracted from questions, they are rich and varied in both style and language of expression. Questions include those directed at cuisine preferences, capacity and age-group constraints, celebrations etc. 
41 \% of the questions contain constraints specifying location requirements (eg near a particular entity). Budgetary and Temporal constraints such as those based on time of day, event in a calendar etc occur in 23\% and 21\% of the questions.

\subsection{Hyper-parameter Settings: } \label{sec:hyper}
For all experiments we set $\delta=1$ in our max-margin criterion. We used Adam Optimizer \cite{Adam} with a learning rate of $0.001$ for training. The convolution layers in the Duet model (retriever) used kernel sizes of $1$ and $3$ for local and distributed interactions respectively. Hidden nodes were initialized with size of input word embeddings, $128$ dimensions. 
The reasoning network (re-ranker) was trained for 5 days on $6$ K80 GPUs (approx. 14 epochs models) using $10$ negative samples for each QA pair. We used $3$-layer $128$-dimensional bidirectional GRUs to encode questions and review sentences. Input word embeddings were updated during training and USE embeddings returned $512$ dimension embeddings.  Training the reasoning network (re-ranker) took $11.5$ hours per epoch on 4 K-80 GPUs. The \csr\ model is trained on negative samples from the a simpler version of the re-ranker with curriculum learning (See Supplementary notes Section \ref{sec:curr}). 

 \subsection{QA System: Answering Characteristics} \label{sec:answer-characteristics}
 
 \begin{figure*}[ht]
 {\footnotesize
 \center
   \includegraphics[scale=0.7]{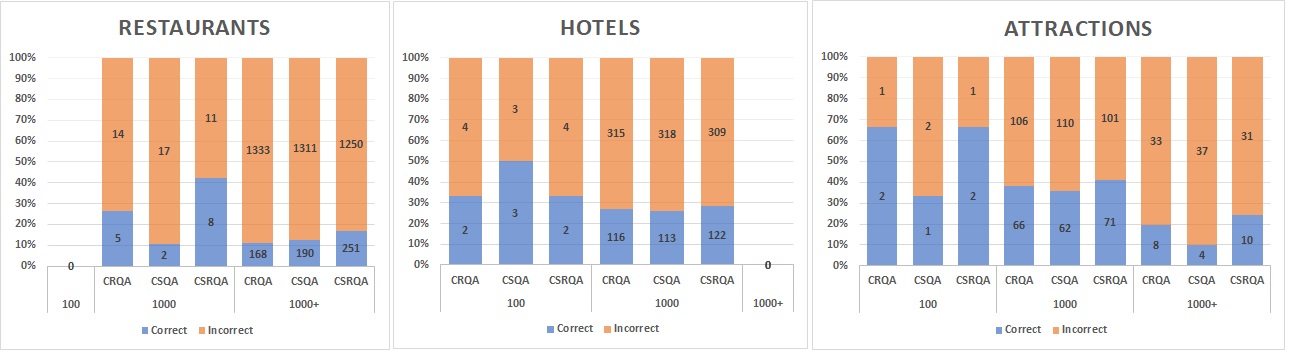}
   \caption{Entity class-wise break-up of the number of times (and \%) a correct answer was within the top-3 ranks binned based on the size of candidate search space (X-axis).}
   \label{fig:space-study}
 
 }
 \end{figure*} 

 We study the performance of different configurations presented in the main paper and their characteristics for each entity-class. 
The plots in Figure \ref{fig:space-study} show the number of times the gold answer was in the top-3 ranks for questions from each entity class\footnote{Recall that each question has its own candidate space}. The results have been binned based on the size of the candidate space (0-100, 100-1000, 1000+). Questions on restaurants dominate the dataset and also have a larger candidate space with $1,501$ questions in the test set having a search space greater than $1,000$ candidates. In this sub-class of questions, we find that the $\cs$ model, which does not do deep reasoning, answers more questions correctly in the top-3 ranks, as compared to the $\crq$ model. This observation strengthens our motivation for using a scalable retrieval model to prune the large search space. 

We find that in hotels and attractions since the search space in most questions isn't as large , both the \cs\ and \crq\ models have comparable performance. However, using the full \csr\ model still shows considerable improvement (8\% relative gain). Overall, we find that the reduction of search space is critical for this task and the use of a scalable shallow neural model to reduce the search space is an effective strategy to improve performance.

 \subsection{QA System: Effect of re-ranking space}
 \label{sec:re-rank}
 
 The performance improvement of the $\csr$ model over the $\crq$ model suggests the re-ranker is easily confused as the set of candidate entities increases. We study the performance of the $\csr$ model by varying the number of candidates it has to re-rank. As expected, as we increase the number of candidates available for re-ranking, the Accuracy@3 begins to drop finally settling at approximately 15\% when the full candidate space is available. However, we find that the drop in  Accuracy@30 increases isn't much suggesting that there are only a few candidates ( approx. 30-40) that the model is confused about. If we had a method of identifying confusing candidates perhaps our model could do better. We test this hypothesis in the next section by experimenting with  different strategies for negative sampling, i.e for sampling {\em harder} candidates for learning the ranker. 
 
 \begin{table}[]
 \small 
 \center
\begin{tabular}{|c||l|l|l|l|}
\hline
\multicolumn{1}{|l||}{\textbf{top-k}} & \textbf{Acc@3} & \textbf{Acc@5} & \textbf{Acc@30} & \textbf{MRR} \\ \hline \hline
10                                 & 19.39          & 25.86          & 33.88           & 0.160        \\ \hline
20                                 & 19.53          & 26.85          & 47.33           & 0.171        \\ \hline
30                                 & 19.01          & 26.66          & 54.32           & 0.171        \\ \hline
40                                 & 18.59          & 26.76          & 57.24           & 0.172        \\ \hline
50                                 & 18.68          & 26.85           & 57.95           & 0.171        \\ \hline
60  & 18.64        &  25.77              &  58.66                              & 0.169             \\ \hline
80             & 18.26                &    25.34            &    58.94             &       0.169       \\ \hline
100           &     18.26           &      25.02          &    58.75             &     0.167         \\ \hline
Full                               & 14.67          & 21.43          & 53.56           & 0.147      \\
\hline
\end{tabular}
\caption{Performance of $\csr$ on the validation data reduces as the candidate space (selected by Duet) to re-rank increases.}
\end{table}

\subsection{Curriculum Learning \& Sampling Strategies for Improved Re-ranking} 
\label{sec:curr}

Max-margin ranking models can be sensitive to the quality of negative samples presented to the model while training. 
Instead of presenting negative samples chosen at random from the candidate space, can we exploit knowledge about entities to give harder samples to the model and help improve its learning? 
One method of selecting harder samples would be to use the gold entity and find entities {\em similar} to it in some latent space and then present the closest entities as negative samples.  In neural settings, candidate embedding space serves as a natural choice for the latent space; negative samples could be generated by sampling from a probability distribution fitted over the distances from the answer embedding.

We also experiment with two baseline methods of creating entity embeddings: (i) Using the averaged sentence embeddings of the representative documents (AVG. Emb in  Table \ref{tab:curr-style}) (ii) Doc2Vec \cite{doc2vec}. We employ curriculum learning, slowly increasing the selection probability of hard negatives up to a maximum of $0.6$. 

One could also use task specific embeddings from $\crq$ to model the candidate space, 
however, running running our trained model on the the test data takes $2.5$\footnote{Using $4$ K-$80$ GPUs} days. Generating question specific candidate embeddings for each instance while training (which is nearly $10$ times larger) is thus infeasible.   
We therefore, decide to generate task specific embeddings using our $\crq$ model but without the Question-Entity-Attention (QEA) layer that learns question independent entity embeddings. Once a model is trained, embeddings can be generated offline and used to generate a probability distribution (per answer entity) for negative sampling.

As can be seen in the last row of Table \ref{tab:curr-style}, training $\crq$ with harder negative samples with curriculum learning helps train a better model.  Interestingly, the negative samples from Duet \cite{DUET,DUET2} results in comparable performance but using Duet as {\em selection} mechanism results in significantly improved performance as shown in the main paper. The $\crq$ model described in the main paper using the task specific embeddings (last row) for training.  

\begin{table}[]
\scriptsize
\begin{tabular}{c||c|c|c|c}
\textbf{Method}                                              & \textbf{Acc@3(\%)} & \textbf{Acc@5(\%)} & \textbf{Acc@30 (\%)} & \textbf{MRR} \\ \hline \hline
\textbf{\begin{tabular}[c]{@{}c@{}}$\crq$ (No CL)\\  \end{tabular}}  &  {16.06}               &  {22.18}               &            {53.04 }     &     {0.155}            \\ \hline
\textbf{\begin{tabular}[c]{@{}c@{}}$\crq$ (CL)\\ Doc2Vec Emb. \end{tabular}} &      16.38     &  22.14              &   {\bf 52.97}                            &    0.149          \\ \hline
\textbf{\begin{tabular}[c]{@{}c@{}}$\crq$ (CL)\\ AVG Emb. \end{tabular}} &  16.24        & 22.14            & 51.68                                & 0.157             \\ \hline
\textbf{\begin{tabular}[c]{@{}c@{}}$\crq$ (CL)\\ Duet Ans. \end{tabular}}   &  16.06  &            21.95   &     52.88  & 0.155   \\  \hline
\textbf{\begin{tabular}[c]{@{}c@{}}$\crq$ (CL)\\ Task Emb. \end{tabular}} &{\bf 16.89}            &  {\bf 23.75}         &            {52.51}   &     {\bf 0.159}    \\   \\ \hline

\end{tabular}
\caption{Curriculum learning (CL) with different entity embedding schemes (Full-ranking task)} \label{tab:curr-style}
\vspace{-2ex}
\end{table}
 
 \subsection{Correlations between Human and Machine relevance judgements} \label{supp:corr}

In order to assess whether performance improvements measured using our gold-data correlate with human judgements on this task, we compute the Spearman's rank coefficient\footnote{https://en.wikipedia.org/wiki/Spearman\%27s\_rank\\\_correlation\_coefficient} between the human assigned $Acc@N$ scores and machine-evaluated $Acc@N$ scores. Let the scoring schemes corresponding to the human and machine judgements be $s_h$, $s_m$ respectively. Let $m_1$ and $m_2$ be any two models developed for our task and let $i^m_{{s}}$ denote the the $Acc@N$ of an question-answer instance $i$ returned by model $m$ $\in$ \{$m_1$,$m_2$\} using scoring scheme $s$ $\in$ \{$s_h$,$s_m$\}.  We then define a random variable $X_s(m_1,m_2)$ as the following: for each question-answer instance $i$, $x^i_s$  is assigned a value of $-1$, $0$ or $1$ based on whether the $Acc@N$ of $m_1$ (according to scoring scheme $s$) is less than, equal to or greater than the $Acc@N$ of $m_2$ as measured under the same scoring scheme. Formally,
$$
x^i_s(m_1,m_2)=
\begin{cases}
-1,\text{if } i^{m_1}_{{s}} $<$ i^{m_2}_{{s}}   \\
0, \text{if } i^{m_1}_{{s}} $=$ i^{m_2}_{{s}} \\
1, \text{if }i^{m_1}_{{s}} $>$ i^{m_2}_{{s}}
\end{cases}
$$

We can now compute the Spearman's rank correlation coefficient $\rho(X_{s_h}(m_1,m_2), X_{s_m}(m_1,m_2)$) using different models. 

Table \ref{tab:spearman} summarizes the correlation coefficients measured between different model pairs. We also report the p-values between each pair which indicates the probability of an uncorrelated system producing data that has a $\rho$ at least as high as the correlation coefficient computed on our data. As can be seen we find there is moderately positive correlation \cite{correlation} with high confidence between the human judgements and gold-data based measurements for both Acc@3 ($\bar{\rho}=0.39$, p-value<0.0009) as well as on Acc@5 ($\bar{\rho}=0.32$ p-value<0.04).  





\begin{table}[]
\scriptsize
\center
\begin{tabular}{|l|l|l|l|l|}
\hline
$m_1$ & $m_2$ & Acc@N & $\rho$ & p-value \\
\hline
\csr   & \cs  & $Acc@3$  & 0.42    &  0.00002       \\
\csr   & \crq & $Acc@3$   & 0.33    &   0.0009
    \\
\cs    & \crq  & $Acc@3$   & 0.43   & 0.000014 \\
\hline
\csr   & \cs  & $Acc@5$  & 0.21    &  0.038
      \\
\csr   & \crq & $Acc@5$   & 0.43    &   0.00001
    \\
\cs    & \crq  & $Acc@5$   & 0.31   & 0.002
 \\
   \hline
\end{tabular}
\caption{Spearman's rank correlation coefficient $\rho$ between human and machine judgements using a pair-wise comparison between different models} 
\label{tab:spearman}
\end{table}

\subsection{Answer System Error Analysis} \label{sec:answer-error}

\begin{figure}[ht]
 {\footnotesize
 \center
  \includegraphics[scale=1.0]{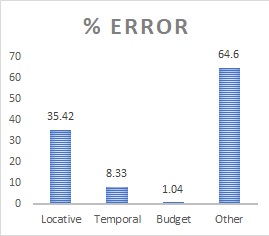}
  \caption{Classification of Errors made by the $\csr$ system (does not sum to 100 because an incorrect answer may exhibit more than one class of errors).}
  \label{fig:answering-analysis}
 
 }
 \end{figure} 

Figure \ref{fig:answering-analysis} gives a detailed break-up of the types of errors made by the $\csr$ system. As can be seen a large set of the errors (35\%) can be traced to answers not fulfilling locative constraints specified in the question. Questions with Budgetary and temporal constraints constitute approximately 9\% of the errors while remaining 65\% of the errors collectively constitute not fulfilling user preferences of cuisine, age appropriate and/or celebration activities, hotel preferences etc. 

\subsection{Dataset Generation} \label{sec:datagen}
We release scripts that regenerate the dataset consisting of the following:
\begin{itemize}
    \item {\bf QA Pairs} containing a question and the set of answer entity IDs. In case of the validation and test data, these question-entity pairs are those generated after crowd-sourced verification. Thus, users of our scripts do not need to run any additional processing apart from executing the crawl scripts. 
    \item {\bf Entity Reviews} for each city mapped by a unique entity ID. 
\end{itemize}

 
Additionally, we also release scripts that allow users to generate new QA pairs. The scripts are organized into the following three stages:
\begin{itemize}
    \item {\bf Crawl:} Download questions and forum post threads based on a seed-url for a city. Additionally, download entities for each city based on a seed-url. We release seed-urls for all our data and those can be used as reference for constructing urls for new data.
    \item { \bf Organize:} Organize the crawled into city specific folders as well generate the silver QA pairs using the entity data as described in Section 3.1 of the main paper.
    \item {\bf Process:} Generate gold QA pairs after executing the high precision extraction steps as described in Section 3.2 of the Main Paper.
\end{itemize}

\subsection{Additional Data statistics} \label{sec:data-stats}
\begin{itemize}
    \item Table \ref{tab:city_Wise_entities} presents the distribution of entities based on their type and the length of review documents for each city. \item As noted in the main paper, most of the cities have restaurants as their majority entity. \item The number of tokens in entity review documents has a huge variance across cities ranging anywhere between $370$ to $8668$. The average length of the questions ranges between 40 - 90 based on the city.
    \item We excluded $3$ cities out of the data set while curating the train, test and validation splits for future studies.
\item Tables \ref{tab:city_Wise_training_qa}, \ref{tab:city_Wise_test_qa}, and \ref{tab:city_Wise_validation_qa} present the city wise distribution of questions and QA Pairs in train, test and validation splits.
\item Restaurant is the most common entity class. 

\end{itemize}

\begin{table*}[h]
\centering
\scriptsize
\begin{tabular}{|l|l|l|l|l|l|l|l|}
\hline
City ID        & \#Attractions & \#Restaurants & \#Hotels & Total Entities & Avg \#Reviews & Avg \#Tokens & \begin{tabular}[c]{@{}l@{}}Avg \#Tokens\\ per Review\end{tabular} \\ \hline
New York       & 846           & 8336          & 562      & 9744           & 83.16         & 4570.5       & 54.9                    \\ \hline
Washington     & 351           & 2213          & 220      & 2784           & 100.7         & 5403.2       & 53.6                    \\ \hline
Chicago        & 471           & 5287          & 174      & 5932           & 51.44         & 2833.4       & 55.1                    \\ \hline
San Francisco  & 426           & 3661          & 302      & 4389           & 60.36         & 3170.6       & 52.5                    \\ \hline
Mexico City    & 290           & 2607          & 318      & 3215           & 26.6          & 1173.5       & 44.1                    \\ \hline
Miami          & 168           & 2283          & 191      & 2642           & 52.5          & 2416.5       & 46.0                    \\ \hline
Vancouver      & 243           & 2518          & 118      & 2879           & 63.17         & 3183.9       & 50.4                    \\ \hline
Sao Paulo      & 248           & 3336          & 232      & 3816           & 9.1           & 370.0        & 40.7                    \\ \hline
Buenos Aires   & 324           & 2385          & 334      & 3043           & 27.17         & 1283.4       & 47.2                    \\ \hline
Rio De Janeiro & 290           & 2320          & 205      & 2815           & 24.54         & 1118.14      & 45.5                    \\ \hline
London         & 1466          & 16212         & 710      & 18388          & 130.46        & 7243.7       & 55.5                    \\ \hline
Dublin         & 387           & 1938          & 270      & 2595           & 160.6         & 8667.8       & 53.9                    \\ \hline
Paris          & 767           & 11379         & 711      & 12857          & 58.6          & 3172.4       & 54.1                    \\ \hline
Rome           & 850           & 6393          & 402      & 7645           & 77.546        & 4115.2       & 53.                     \\ \hline
Stockholm      & 200           & 2168          & 125      & 2493           & 56.39         & 2646.8       & 46.9                    \\ \hline
Oslo           & 211           & 1061          & 74       & 1346           & 73.06         & 3362.0       & 46.0                    \\ \hline
Zurich         & 144           & 1434          & 97       & 1675           & 47.81         & 2162.0       & 45.2                    \\ \hline
Vienna         & 412           & 2761          & 332      & 3505           & 82.33         & 3724.7       & 45.2                    \\ \hline
Berlin         & 518           & 5147          & 593      & 6258           & 64.99         & 2958.4       & 45.5                    \\ \hline
Budapest       & 340           & 2225          & 170      & 2735           & 123.26        & 5762.8       & 46.7                    \\ \hline
Bucharest      & 212           & 1424          & 196      & 1832           & 47.76         & 2043.1       & 42.7                    \\ \hline
Moscow         & 544           & 3291          & 259      & 4094           & 21.73         & 946.1        & 43.5                    \\ \hline
Amsterdam      & 358           & 3055          & 422      & 3835           & 116.8         & 5769.9       & 49.4                    \\ \hline
Beijing        & 509           & 2234          & 0        & 2743           & 18.87         & 1067.4       & 56.6                    \\ \hline
New Delhi      & 350           & 5102          & 671      & 6123           & 31.72         & 1317.7       & 41.5                    \\ \hline
Mumbai         & 432           & 7159          & 383      & 7974           & 22.45         & 881.6        & 39.3                    \\ \hline
Agra           & 66            & 250           & 203      & 519            & 93.27         & 3979.7       & 42.6                    \\ \hline
Bangkok        & 435           & 5778          & 793      & 7006           & 54.48         & 2457.0       & 45.1                    \\ \hline
Karachi        & 62            & 219           & 14       & 295            & 22.31         & 869.4        & 38.9                    \\ \hline
Singapore      & 28            & 7616          & 453      & 8097           & 42.52         & 1965.2       & 46.2                    \\ \hline
Jakarta        & 194           & 3853          & 602      & 4649           & 25.64         & 887.3        & 34.6                    \\ \hline
Tokyo          & 0             & 0             & 781      & 781            & 157.04        & 6215.8       & 39.6                    \\ \hline
Seoul          & 354           & 3747          & 611      & 4712           & 25.53         & 1098.8       & 43.0                    \\ \hline
Bukhara        & 38            & 23            & 14       & 75             & 24.4          & 1112.3       & 45.5                    \\ \hline
Ulaanbaatar    & 56            & 222           & 47       & 325            & 23.75         & 1103.1       & 46.4                    \\ \hline
Kathmandu      & 111           & 588           & 178      & 877            & 54.07         & 2356.9       & 43.6                    \\ \hline
Melbourne      & 324           & 3030          & 162      & 3516           & 49.31         & 2464.2       & 49.9                    \\ \hline
Sydney         & 353           & 4100          & 362      & 4815           & 64.57         & 3125.5       & 48.4                    \\ \hline
Auckland       & 175           & 1733          & 238      & 2146           & 48.5          & 2330.7       & 48.0                    \\ \hline
Havana         & 183           & 657           & 23       & 863            & 44.07         & 2394.8       & 54.3                    \\ \hline
Honolulu       & 218           & 1561          & 117      & 1896           & 88.2          & 4762.8       & 54.0                    \\ \hline
Kingston       & 39            & 159           & 54       & 252            & 68.6          & 2721.9       & 39.7                    \\ \hline
Seychelles     & 0             & 1             & 0        & 1              & 20.0          & 1306.0       & 65.3                    \\ \hline
Dubai          & 247           & 5786          & 347      & 6380           & 54.14         & 2419.6       & 44.7                    \\ \hline
Cairo          & 155           & 1232          & 111      & 1498           & 33.27         & 1344.9       & 40.4                    \\ \hline
Amman          & 41            & 499           & 143      & 683            & 58.52         & 2264.7       & 38.7                    \\ \hline
Jerusalem      & 227           & 561           & 31       & 819            & 55.93         & 2651.8       & 47.4                    \\ \hline
Johannesburg   & 111           & 929           & 180      & 1220           & 58.559        & 2262.5       & 38.6                    \\ \hline
Cape Town      & 139           & 822           & 287      & 1248           & 121.638       & 4969.0       & 40.8                    \\ \hline
Nairobi        & 72            & 477           & 107      & 656            & 42.2698       & 2013.7       & 47.6\\ \hline            
\end{tabular}
\caption{City Wise -  Knowledge Source Statistics} \label{tab:city_Wise_entities}
\end{table*}

\begin{table*}[]
\centering
\scriptsize
\begin{tabular}{|l|l|l|l|l|l|l|}
\hline
City Name      & \#Questions & \#QA Pairs & \begin{tabular}[c]{@{}l@{}}\#QA Pairs\\ With Hotel\end{tabular} & \begin{tabular}[c]{@{}l@{}}\#QA Pairs\\ With Restaurants\end{tabular} & \begin{tabular}[c]{@{}l@{}}\#QA Pairs\\ With Attractions\end{tabular} & \begin{tabular}[c]{@{}l@{}}Avg \#Tokens\\ in Question\end{tabular} \\ \hline
New York      & 5891 & 14673 & 1030 & 12841 & 802  & 77.1 \\ \hline
Washington    & 861  & 1886  & 168  & 1591  & 127  & 73.2 \\ \hline
Chicago       & 1189 & 2888  & 129  & 2583  & 176  & 76.2 \\ \hline
San Francisco & 1621 & 4079  & 410  & 3417  & 252  & 74.0 \\ \hline
Mexico City   & 127  & 216   & 65   & 137   & 14   & 68.4 \\ \hline
Miami         & 98   & 134   & 28   & 97    & 9    & 68.2 \\ \hline
Vancouver     & 498  & 874   & 223  & 554   & 97   & 74.9 \\ \hline
Sao Paulo     & 16   & 25    & 7    & 16    & 2    & 75.7 \\ \hline
Buenos Aires  & 268  & 493   & 140  & 325   & 28   & 77.2 \\ \hline
London        & 3387 & 8265  & 569  & 6572  & 1124 & 75.9 \\ \hline
Dublin        & 621  & 1103  & 196  & 810   & 97   & 72.5 \\ \hline
Rome          & 1004 & 1782  & 234  & 1292  & 256  & 72.4 \\ \hline
Stockholm     & 160  & 280   & 56   & 190   & 34   & 78.1 \\ \hline
Oslo          & 67   & 114   & 43   & 65    & 6    & 78.2 \\ \hline
Zurich        & 95   & 147   & 41   & 97    & 9    & 69.8 \\ \hline
Vienna        & 292  & 465   & 89   & 320   & 56   & 66.0 \\ \hline
Berlin        & 386  & 652   & 68   & 453   & 131  & 71.8 \\ \hline
Budapest      & 317  & 655   & 23   & 605   & 27   & 75.3 \\ \hline
Bucharest     & 22   & 46    & 3    & 41    & 2    & 59.8 \\ \hline
Moscow        & 64   & 106   & 26   & 74    & 6    & 70.2 \\ \hline
Amsterdam     & 669  & 1299  & 207  & 1002  & 90   & 70.6 \\ \hline
Beijing       & 54   & 71    & 0    & 57    & 14   & 70.7 \\ \hline
New Delhi     & 28   & 55    & 24   & 18    & 13   & 54.0 \\ \hline
Mumbai        & 166  & 334   & 98   & 198   & 38   & 63.8 \\ \hline
Agra          & 40   & 52    & 36   & 14    & 2    & 54.8 \\ \hline
Bangkok       & 743  & 963   & 313  & 482   & 168  & 65.1 \\ \hline
Singapore     & 515  & 821   & 332  & 471   & 18   & 67.6 \\ \hline
Jakarta       & 25   & 44    & 15   & 15    & 14   & 64.7 \\ \hline
Tokyo         & 16   & 22    & 22   & 0     & 0    & 64.6 \\ \hline
Seoul         & 70   & 82    & 39   & 29    & 14   & 69.4 \\ \hline
Kathmandu     & 23   & 39    & 26   & 13    & 0    & 75.7 \\ \hline
Melbourne     & 33   & 65    & 5    & 56    & 4    & 63.7 \\ \hline
Sydney        & 344  & 508   & 100  & 340   & 68   & 67.0 \\ \hline
Havana        & 37   & 52    & 8    & 39    & 5    & 70.0 \\ \hline
Honolulu      & 61   & 93    & 24   & 61    & 8    & 56.0 \\ \hline
Kingston      & 5    & 6     & 1    & 4     & 1    & 87.6 \\ \hline
Cairo         & 48   & 57    & 10   & 36    & 11   & 77.8 \\ \hline
Amman         & 9    & 10    & 3    & 7     & 0    & 56.7 \\ \hline
Jerusalem     & 44   & 58    & 4    & 43    & 11   & 57.9 \\ \hline
Johannesburg  & 17   & 20    & 14   & 4     & 2    & 66.3 \\ \hline
Cape Town     & 40   & 57    & 26   & 31    & 0    & 65.3 \\ \hline
Nairobi       & 25   & 28    & 5    & 19    & 4    & 66.3 \\ \hline
\end{tabular}
\caption{City Wise Training Dataset Statistics} \label{tab:city_Wise_training_qa}
\end{table*}

\begin{table*}[]
\centering
\scriptsize
\begin{tabular}{|l|l|l|l|l|l|l|}
\hline
City Name      & \#Questions & \#QA Pairs & \begin{tabular}[c]{@{}l@{}}\#QA Pairs\\ With Hotel\end{tabular} & \begin{tabular}[c]{@{}l@{}}\#QA Pairs\\ With Restaurants\end{tabular} & \begin{tabular}[c]{@{}l@{}}\#QA Pairs\\ With Attractions\end{tabular} & \begin{tabular}[c]{@{}l@{}}Avg \#Tokens\\ in Question\end{tabular} \\ \hline
New York &  627 & 1445 & 116 &1243 & 86 & 77.0 \\ \hline 
Washington &  104 & 243 & 18 &213 & 12 & 80.9 \\ \hline 
Chicago &  141 & 324 & 16 &295 & 13 & 74.2 \\ \hline 
San Francisco &  185 & 439 & 38 &360 & 41 & 73.9 \\ \hline 
Mexico City &  14 & 20 & 7 &9 & 4 & 62.4 \\ \hline 
Miami &  13 & 16 & 2 &9 & 5 & 54.7 \\ \hline 
Vancouver &  53 & 99 & 26 &57 & 16 & 74.3 \\ \hline 
Sao Paulo &  1 & 1 & 1 &0 & 0 & 65.0 \\ \hline 
Buenos Aires &  39 & 82 & 15 &66 & 1 & 68.7 \\ \hline 
London &  342 & 634 & 76 &469 & 89 & 75.2 \\ \hline 
Dublin &  62 & 122 & 20 &97 & 5 & 76.9 \\ \hline 
Rome &  118 & 185 & 25 &139 & 21 & 72.7 \\ \hline 
Stockholm &  24 & 46 & 9 &29 & 8 & 82.6 \\ \hline 
Oslo &  9 & 12 & 5 &7 & 0 & 82.9 \\ \hline 
Zurich &  11 & 16 & 3 &13 & 0 & 53.4 \\ \hline 
Vienna &  29 & 47 & 9 &32 & 6 & 55.7 \\ \hline 
Berlin &  39 & 60 & 13 &37 & 10 & 82.6 \\ \hline 
Budapest &  48 & 96 & 3 &87 & 6 & 66.5 \\ \hline 
Bucharest &  2 & 7 & 0 &7 & 0 & 44.5 \\ \hline 
Moscow &  9 & 14 & 7 &5 & 2 & 65.7 \\ \hline 
Amsterdam &  72 & 113 & 30 &74 & 9 & 75.5 \\ \hline 
Beijing &  7 & 8 & 0 &7 & 1 & 65.9 \\ \hline 
New Delhi &  1 & 3 & 0 &3 & 0 & 75.0 \\ \hline 
Mumbai &  20 & 38 & 20 &14 & 4 & 64.0 \\ \hline 
Agra &  4 & 7 & 5 &2 & 0 & 37.5 \\ \hline 
Bangkok &  56 & 68 & 32 &26 & 10 & 64.8 \\ \hline 
Singapore &  46 & 72 & 30 &41 & 1 & 62.2 \\ \hline 
Jakarta &  3 & 4 & 3 &1 & 0 & 43.3 \\ \hline 
Tokyo &  1 & 1 & 1 &0 & 0 & 42.0 \\ \hline 
Seoul &  4 & 4 & 3 &1 & 0 & 50.8 \\ \hline 
Kathmandu &  1 & 3 & 3 &0 & 0 & 12.0 \\ \hline 
Melbourne &  2 & 2 & 0 &0 & 2 & 31.5 \\ \hline 
Sydney &  39 & 50 & 12 &32 & 6 & 80.4 \\ \hline 
Havana &  3 & 4 & 0 &3 & 1 & 75.0 \\ \hline 
Honolulu &  8 & 8 & 2 &5 & 1 & 73.2 \\ \hline 
Kingston &  1 & 1 & 0 &0 & 1 & 78.0 \\ \hline 
Cairo &  12 & 15 & 0 &14 & 1 & 77.8 \\ \hline 
Amman &  3 & 4 & 0 &3 & 1 & 83.3 \\ \hline 
Jerusalem &  6 & 8 & 0 &5 & 3 & 87.3 \\ \hline 
Johannesburg &  4 & 5 & 1 &4 & 0 & 47.5 \\ \hline 
Cape Town &  7 & 13 & 6 &7 & 0 & 70.6 \\ \hline 
Nairobi &  3 & 3 & 1 &2 & 0 & 68.3 \\ \hline 

\end{tabular}
\caption{City Wise Test Dataset Statistics} \label{tab:city_Wise_test_qa}
\end{table*}

\begin{table*}[]
\centering
\scriptsize
\begin{tabular}{|l|l|l|l|l|l|l|}
\hline
City Name      & \#Questions & \#QA Pairs & \begin{tabular}[c]{@{}l@{}}\#QA Pairs\\ With Hotel\end{tabular} & \begin{tabular}[c]{@{}l@{}}\#QA Pairs\\ With Restaurants\end{tabular} & \begin{tabular}[c]{@{}l@{}}\#QA Pairs\\ With Attractions\end{tabular} & \begin{tabular}[c]{@{}l@{}}Avg \#Tokens\\ in Question\end{tabular} \\ \hline
New York &  621 & 1362 & 119 &1169 & 74 & 75.6 \\ \hline 
Washington &  114 & 236 & 20 &202 & 14 & 74.0 \\ \hline 
Chicago &  140 & 334 & 20 &293 & 21 & 71.6 \\ \hline 
San Francisco &  171 & 413 & 55 &328 & 30 & 78.2 \\ \hline 
Mexico City &  16 & 20 & 10 &8 & 2 & 77.3 \\ \hline 
Miami &  7 & 8 & 3 &5 & 0 & 60.6 \\ \hline 
Vancouver &  61 & 102 & 27 &65 & 10 & 74.2 \\ \hline 
Sao Paulo &  3 & 8 & 2 &6 & 0 & 83.7 \\ \hline 
Buenos Aires &  25 & 46 & 13 &33 & 0 & 79.5 \\ \hline 
London &  334 & 657 & 81 &494 & 82 & 74.5 \\ \hline 
Dublin &  71 & 125 & 34 &85 & 6 & 72.5 \\ \hline 
Rome &  108 & 166 & 25 &119 & 22 & 71.1 \\ \hline 
Stockholm &  17 & 32 & 7 &18 & 7 & 62.9 \\ \hline 
Oslo &  8 & 9 & 5 &4 & 0 & 78.5 \\ \hline 
Zurich &  17 & 26 & 12 &14 & 0 & 73.2 \\ \hline 
Vienna &  37 & 59 & 12 &36 & 11 & 72.8 \\ \hline 
Berlin &  28 & 46 & 12 &28 & 6 & 73.8 \\ \hline 
Budapest &  34 & 58 & 3 &54 & 1 & 66.8 \\ \hline 
Bucharest &  1 & 2 & 2 &0 & 0 & 89.0 \\ \hline 
Moscow &  6 & 14 & 4 &10 & 0 & 57.3 \\ \hline 
Amsterdam &  72 & 140 & 11 &121 & 8 & 72.2 \\ \hline 
Beijing &  3 & 5 & 0 &4 & 1 & 36.0 \\ \hline 
New Delhi &  5 & 6 & 1 &3 & 2 & 24.2 \\ \hline 
Mumbai &  15 & 32 & 9 &21 & 2 & 71.5 \\ \hline 
Agra &  3 & 5 & 4 &1 & 0 & 33.3 \\ \hline 
Bangkok &  55 & 71 & 26 &32 & 13 & 66.3 \\ \hline 
Karachi &  1 & 1 & 0 &0 & 1 & 78.0 \\ \hline 
Singapore &  53 & 81 & 37 &42 & 2 & 69.1 \\ \hline 
Jakarta &  3 & 8 & 5 &3 & 0 & 55.7 \\ \hline 
Seoul &  8 & 8 & 6 &0 & 2 & 57.9 \\ \hline 
Kathmandu &  4 & 6 & 6 &0 & 0 & 86.8 \\ \hline 
Melbourne &  2 & 4 & 0 &4 & 0 & 74.5 \\ \hline 
Sydney &  35 & 56 & 4 &44 & 8 & 62.0 \\ \hline 
Havana &  4 & 5 & 1 &4 & 0 & 72.8 \\ \hline 
Honolulu &  13 & 15 & 3 &11 & 1 & 62.2 \\ \hline 
Cairo &  8 & 13 & 2 &7 & 4 & 64.4 \\ \hline 
Jerusalem &  6 & 6 & 0 &5 & 1 & 40.5 \\ \hline 
Johannesburg &  3 & 3 & 1 &1 & 1 & 77.7 \\ \hline 
Cape Town &  4 & 5 & 1 &2 & 2 & 71.2 \\ \hline 
Nairobi &  3 & 3 & 2 &0 & 1 & 81.7 \\ \hline 

\end{tabular}
\caption{City Wise Validation Dataset Statistics} \label{tab:city_Wise_validation_qa}
\end{table*}



\end{document}